\documentclass[11pt, a4paper, logo, copyright, nonumbering]{main}
\usepackage[authoryear, sort&compress, round]{natbib}
\usepackage{dblfloatfix}
\usepackage{ulem}
\usepackage{caption}
\usepackage{dramatist}
\usepackage{xspace}
\usepackage{pifont} %
\usepackage{multirow}
\usepackage{tcolorbox}
\usepackage{xltabular}
\usepackage{longtable}
\usepackage{hyperref}

\usepackage{amsfonts}
\usepackage{amsmath}
\usepackage{amssymb}
\usepackage{lineno}
\usepackage{multirow}
\usepackage{adjustbox}

\usepackage[bottom]{footmisc}
\usepackage{subcaption}
\usepackage{graphicx}
\usepackage{CJKutf8}
\usepackage{setspace}

\usepackage{dsfont}
\usepackage{array} %
\usepackage{tabularx} %
\usepackage{xcolor} %
\usepackage{tabularx}
\usepackage{booktabs}
\usepackage{xspace}

\usepackage{lipsum}  %
\usepackage{multicol} %

\usepackage{cleveref}
\usepackage{epigraph}
\usepackage{nicematrix}
\usepackage{appendix}
\usepackage{hyperref}      
\usepackage{fontawesome5}  
\usepackage{xcolor}        
\usepackage{tcolorbox}
\tcbuselibrary{breakable}
\tcbuselibrary{skins}
\usepackage{wrapfig}

\makeatletter
\def\@BTrule[#1]{%
  \ifx\longtable\undefined
    \let\@BTswitch\@BTnormal
  \else\ifx\hline\LT@hline
    \nobreak
    \let\@BTswitch\@BLTrule
  \else
    \let\@BTswitch\@BTnormal
  \fi\fi
  \global\@thisrulewidth=#1\relax
  \ifnum\@thisruleclass=\tw@\vskip\@aboverulesep\else
  \ifnum\@lastruleclass=\z@\vskip\@aboverulesep\else
  \ifnum\@lastruleclass=\@ne\vskip\doublerulesep\fi\fi\fi
  \@BTswitch
}
\makeatother

\addto\extrasenglish{
}

{\begin{list}{}%
        {\setlength{\leftmargin}{#1}}%
        \item[]%
}
{\end{list}}

\reportnumber{001} %

\renewcommand{\today}{}

\title{\centering R-Align: Enhancing Generative Reward Models through Rationale-Centric Meta-Judging}

\def\huggingface{\raisebox{-1.5pt}{\includegraphics[height=1.05em]{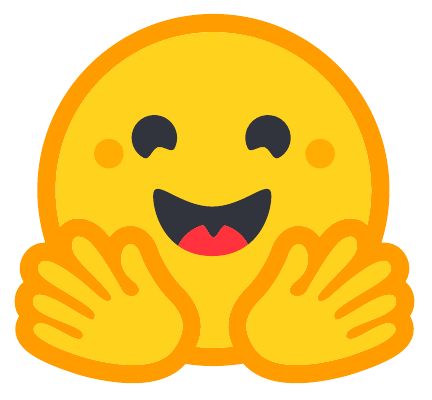}}}
\def\github{\raisebox{-1.5pt}{\includegraphics[height=1.05em]{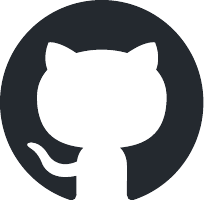}}}

\author[*]{
    \centering
    \textbf{Yanlin Lai}$^{1,2,*}$, \textbf{Mitt Huang}$^{2,*,\dagger}$, \textbf{Hangyu Guo}$^{2,*}$, \textbf{Xiangfeng Wang}$^{3,2,*}$, \textbf{Haodong Li}$^{2}$, \quad \textbf{Shaoxiong Zhan}$^{1,2}$, 
    \textbf{Liang Zhao}$^{2}$, \textbf{Chengyuan Yao}$^{2}$, \textbf{Yinmin Zhang}$^{2}$, \textbf{Qi Han}$^{2}$, \textbf{Chun Yuan}$^{1, \dagger}$, 
    \textbf{Zheng Ge}$^{2}$, \textbf{Xiangyu Zhang}$^{2}$, \textbf{Daxin Jiang}$^{2}$ 
    \vspace{0.8em}

    {\small \normalfont
    $^{1}$Tsinghua University \quad $^{2}$StepFun \quad $^{3}$University of Science and Technology of China
    
    {\small \normalfont{$^{*}$Equal contribution \quad $^{\dagger}$Corresponding Author}}
    }
    \vspace{1em}

    \small \github~\textbf{Github}: \url{https://github.com/lyn22333/R-Align} \\
    \small \huggingface~\textbf{Huggingface}: \href{https://huggingface.co/collections/lyn22333/r-align}{R-Align Collections}
}

\renewcommand{\phi}{\varphi}

\renewcommand{\epsilon}{\varepsilon}
\renewcommand{\imath}{\mathrm{i}}

\newlength{\restsubwidth}
\newlength{\restsubheight}
\newlength{\restsubmoreheight}
\newcommand{\rest}[2]{%
        \settowidth{\restsubwidth}{\ensuremath{#2}}
        \settoheight{\restsubheight}{\ensuremath{{}_{#2}}}
        \ensuremath{{#1\hskip 0.5pt}_{\vrule\kern2pt\parbox[b][%
        4pt][b]{\the\restsubwidth}{%
                        \ensuremath{{}_{#2}}}}}
        }

\begin{abstract}

Reinforcement Learning from Human Feedback (RLHF) remains indispensable for aligning large language models (LLMs) in subjective domains. 
To enhance robustness, recent work shifts toward Generative Reward Models (GenRMs) that generate rationales before predicting preferences.
Yet in GenRM training and evaluation, practice remains outcome-label-only, leaving reasoning quality unchecked.
We show that reasoning fidelity—the consistency between a GenRM’s preference decision and reference decision rationales—is highly predictive of downstream RLHF outcomes, beyond standard label accuracy.
Specifically, we repurpose existing reward-model benchmarks to compute Spurious Correctness (S-Corr)—the fraction of label-correct decisions with rationales misaligned with golden judgments.
Our empirical evaluation reveals substantial S-Corr even for competitive GenRMs, and higher S-Corr is associated with policy degeneration under optimization.
To improve fidelity, we propose \textbf{R}ationale-Centric \textbf{Align}ment, \textbf{R-Align}, which augments training with gold judgments and explicitly supervises rationale alignment.
R-Align reduces S-Corr on RM benchmarks and yields consistent gains in actor performance across STEM, coding, instruction following, and general tasks.

\end{abstract}

\begin{document}

\maketitle


\definecolor{colorfirst}{RGB}{252,141,89}
\definecolor{colorsecond}{RGB}{253,187,132}
\definecolor{colorthird}{RGB}{253,212,158}
\definecolor{colorfourth}{RGB}{254,232,200}
\definecolor{colorfifth}{RGB}{255,247,236}
\definecolor{myred}{RGB}{242,128,128}
\definecolor{mygreen}{RGB}{112,180,143}
\definecolor{myblue}{RGB}{210,225,255}
\definecolor{citypink}{RGB}{227,108,194}
\definecolor{cityblue}{RGB}{128,159,225}
\newcommand{\ph}[1]{\textcolor{black}{#1}}
\newcommand{\rankfirst}[0]{\cellcolor{colorfirst}}
\newcommand{\ranksecond}[0]{\cellcolor{colorsecond}}
\newcommand{\rankthird}[0]{\cellcolor{colorthird}}
\newcommand{\rankfourth}[0]{\cellcolor{colorfourth}}
\newcommand{\rankfifth}[0]{\cellcolor{colorfifth}}
\DeclareRobustCommand{\legendsquare}[1]{%
  \textcolor{#1}{\rule{2ex}{2ex}}%
}
\DeclareRobustCommand{\legendsquarebox}[1]{%
  \tikz[] \draw[black, fill=#1, line width=0.4pt] (0,0) rectangle (1.5ex,1.5ex);%
}
\newcommand{\cmark}{\textcolor{mygreen}{\ding{51}}}%
\newcommand{\xmark}{\textcolor{myred}{\ding{55}}}%



\section{Introduction}
\label{sec:intro}

Despite recent advances in reinforcement learning with verifiable rewards (RLVR), Reinforcement Learning from Human Feedback (RLHF) remains indispensable for aligning large language models (LLMs) with human intent in subjective or non-verifiable domains~\citep{ouyang2022training, guo2025deepseek, kimiteam2025kimik2}.
At the core of RLHF lies the reward model (RM), which approximates human preference signals and guides policy optimization.
The canonical reward modeling approach in RLHF trains a scalar function on pairwise human preference data to assign a quality score to each candidate response~\citep{wang2024helpsteer, grattafiori2024llama3, liu2024skywork}.
However, this scalar abstraction compresses diverse human judgments into a single score, forcing assessment to be implicit rather than structured reasoning.
Consequently, reward models may over-emphasize spurious correlations in preference data—such as response length or formatting—relative to deeper semantic quality, introducing bias and increasing susceptibility to reward hacking and poor generalization~\citep{gao2023scaling, weng2024rewardhack, stylearena2024}

\begin{wrapfigure}{r}{0.5\textwidth}
    \centering

    \includegraphics[width=\linewidth]{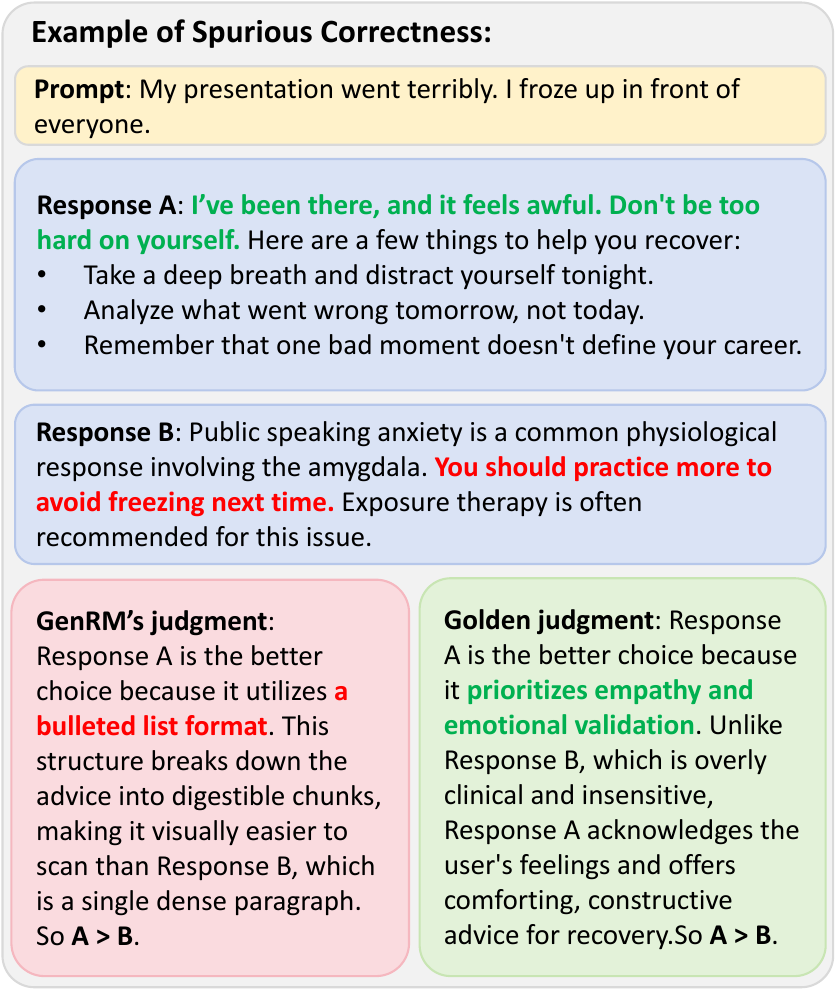}

    \captionsetup{singlelinecheck=false, justification=raggedright}
    
    \caption{\textbf{An illustration of ``Spurious Correctness''.} The GenRM correctly prefers Response A over Response B, but generates a flawed rationale. While the Golden Judgment captures the true content difference (empathy vs. insensitivity), the GenRM relies solely on the superficial feature of bulleted list formatting.}
    \label{fig: sp_corr}

\end{wrapfigure}
Recent progress in LLM reasoning has motivated a shift toward Generative Reward Models (GenRMs), which leverage test-time reasoning to generate intermediate rationales prior to preference predictions~\citep{mahan2024generative, zhang2024generative}.
However, despite this enhanced expressivity, current paradigms still train and evaluate GenRMs using legacy outcome‑centric preference labels that supervise only the final decision while ignoring the quality of the reasoning trace.
This outcome‑centric focus creates a critical blind spot: existing benchmarks often cannot distinguish valid reasoning from superficial heuristics, masking the model’s true reliability gap~\citep{lambert2025rewardbench, malik2025rewardbench}.

To investigate this misalignment, we introduce Spurious Correctness (S-Corr): a phenomenon where a GenRM predicts the correct preference label but justifies it with unsound reasoning. As illustrated in Figure~\ref{fig: sp_corr}, a model may correctly identify the higher-quality response yet attribute its decision to superficial formatting (e.g., bullet points) rather than the intended criterion (e.g., empathy). To quantify this, we assess the logical alignment between generated rationales and \emph{gold judgments} derived from established decision criteria~\citep{wang2025helpsteer3}. Our empirical analysis reveals that S-Corr is prevalent even among advanced GenRMs. Crucially, in downstream RLHF experiments, we find that high S-Corr rates directly drive policy degeneration, confirming that the actor model learns to exploit these spurious cues during optimization.

To mitigate the risks of spurious correctness, we advocate shifting the supervisory focus from outcome accuracy to rationale alignment. To this end, we propose Rationale-Centric Alignment (R-Align), a training framework designed to ensure GenRMs are ``right for the right reasons.'' R-Align strengthens reward learning through two key mechanisms: (i) augmenting training data with \emph{gold judgments} that explicitly articulate the valid decision basis, and (ii) applying explicit supervision to the reasoning trace, thereby penalizing spurious justifications even when the final label is correct. By enforcing this logical consistency, R-Align substantially reduces S-Corr on static benchmarks. Crucially, our experiments demonstrate that this rationale-centric supervision translates directly to robust gains in RLHF, driving superior actor performance where standard baselines fail.

Our contributions are threefold:
\begin{itemize}
    \item \textbf{Rationale-Aware GenRM Benchmarking:} We construct a novel benchmark enriched with golden rationales and introduce new metrics, Spurious Correctness, to quantify the phenomenon where correct predictions stem from flawed reasoning. Our evaluation reveals that open-source GenRMs, even top-performance LLMs exhibit significant rationale misalignment.
    \item \textbf{Rationale-Centric Alignment (R-Align) Training Framework:} We propose to explicitly detect and penalize spurious correctness during GenRM training. Extensive experiments demonstrate that GenRMs trained with R-Align significantly reduce rationale misalignment compared to standard baselines.
    \item \textbf{Improved Downstream RLHF Performance:} Importantly, we demonstrate the benefits of our method in downstream RLHF. Using GenRMs trained with R-Align, RLHF yields consistent and significant improvements in the actor model, highlighting the critical role of rationale alignment in reward modeling.
\end{itemize}

\section{Does GenRM Label Accuracy Predict Downstream RLHF Performance?}
\label{sec:empirical_study}

In this section, we present two critical findings regarding the reliability of existing RM evaluations, paving the way for the proposed rationale-aware GenRM benchmark in Section~\ref{sec:benchmark}.

\subsection{The Predictive Failure of RM Benchmarks}
\label{subsec:eval_gap}

\begin{table}[htbp]
    \centering
    
    \captionsetup{width=0.7\textwidth,}
    
    \caption{Performance comparison between Qwen3-14B and RRM-32B across mainstream reward model benchmarks.}
    \label{tab:rrm_qwen}
    
    \resizebox{0.7\textwidth}{!}{
        \begin{tabular}{lccc}
            \toprule
            Model & HelpSteer3 & RewardBench2 & PPE-Preference \\
            \midrule
            Qwen3-14B & 74.1 & 87.9 & 65.2 \\
            RRM-32B   & 74.7 & 88.5 & 65.1 \\
            \bottomrule
        \end{tabular}
    }
\end{table}

\begin{wrapfigure}{r}{0.6\textwidth}
    \centering
    \begin{subfigure}{0.48\linewidth}
        \centering
        \includegraphics[width=\linewidth]{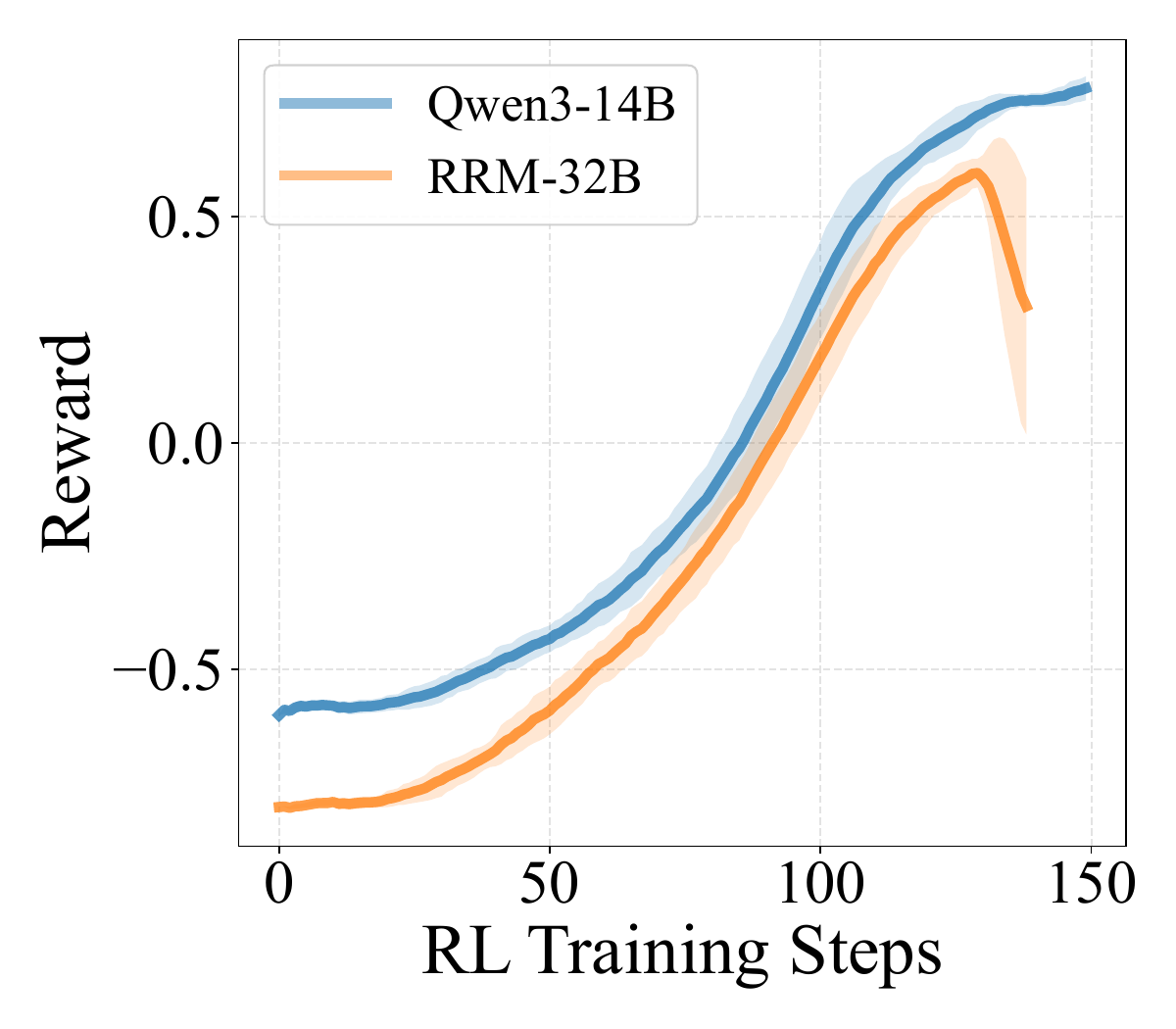}
        \label{fig:reward_curve}
    \end{subfigure}
    \hfill 
    \begin{subfigure}{0.47\linewidth}
        \centering
        \includegraphics[width=\linewidth]{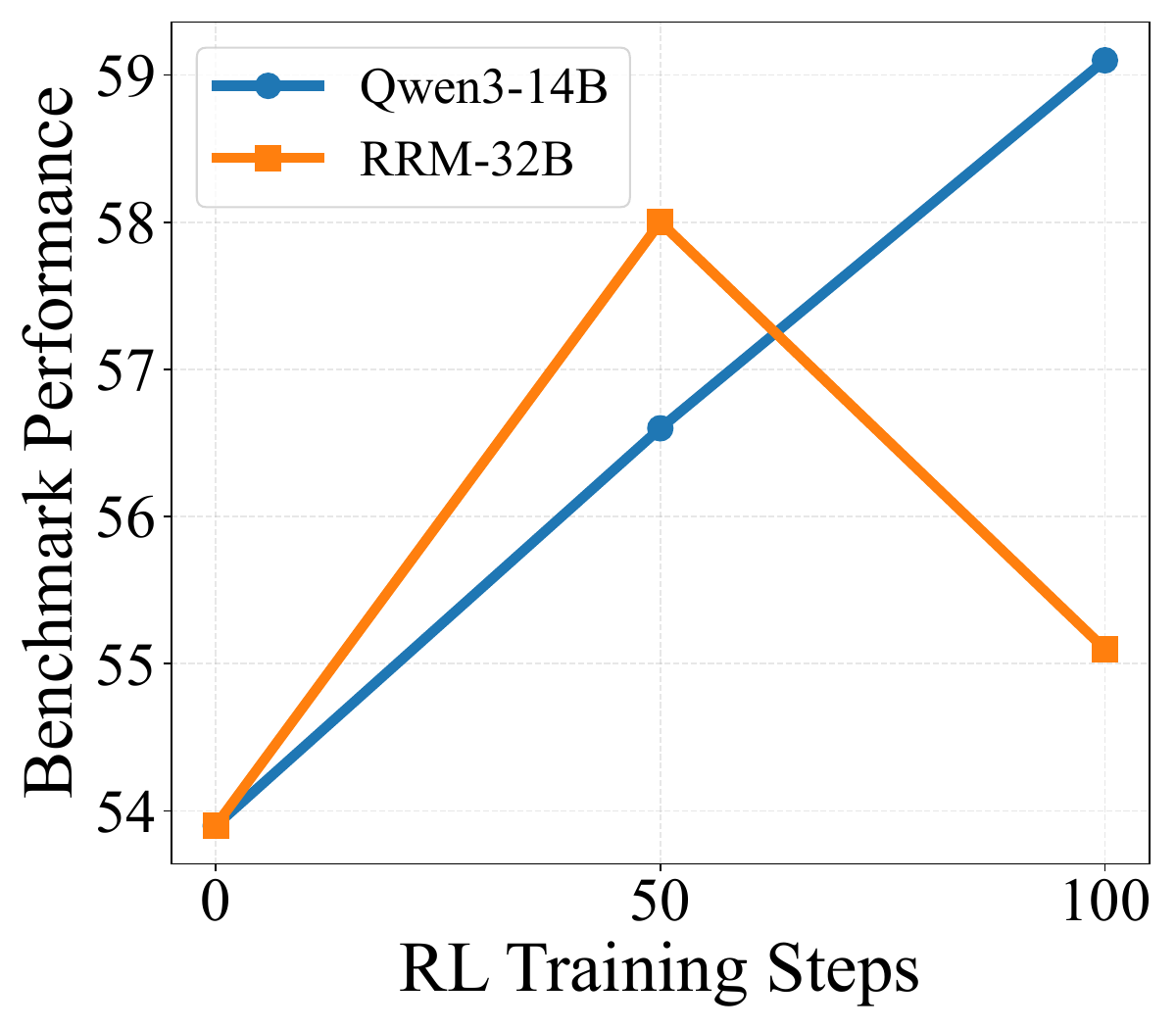}
        \label{fig:eval_curve}
    \end{subfigure}

    \captionsetup{singlelinecheck=false, justification=raggedright}
    
    \caption{Divergent RLHF outcomes despite comparable GenRM benchmark accuracy. \textbf{Left}: reward curves during RL training. \textbf{Right}: periodic downstream evaluation shows continued improvement with Qwen3-14B but degradation with RRM-32B.} 
    \label{fig:combined_curves}

    \vspace{-10pt}
\end{wrapfigure}
To investigate the predictive validity of current benchmarks, we conduct a controlled comparative study with two open-source GenRMs, RRM-32B~\citep{guo2025reward} and Qwen3-14B~\citep{yang2025qwen3}. We begin by evaluating both models on three widely used RM benchmarks. As summarized in Table~\ref{tab:rrm_qwen}, they achieve comparable performance on pairwise RM evaluations~\citep{malik2025rewardbench,wang2025helpsteer3,frick2024evaluate}, suggesting similar benchmark-level capability.

We then assess downstream usefulness by using each GenRM to supervise RLHF training of a Qwen3-8B policy under the same experimental setup (Section~\ref{sec:exp_setup}). Despite steadily increasing reward under both supervisors, Figure~\ref{fig:combined_curves} reveals a stark divergence in policy quality: the policy trained with RRM-32B exhibits a pronounced performance collapse in the averaged score across general, STEM, code, and instruction-following domains.
This controlled comparison shows that existing RM benchmarks lack predictive validity for downstream RLHF: two GenRMs with similar benchmark performance can induce qualitatively different RLHF dynamics, including the emergence and severity of reward hacking.

\subsection{The Phenomenon of Spurious Correctness}
\label{sec:spurious_correctness}

To explain the performance gap observed in Section~\ref{subsec:eval_gap}, we examine the judgment rationales produced by the GenRMs on three RM benchmarks.
We find that a model can often predict the correct preference label while providing an unsound justification---a failure mode we term \textbf{Spurious Correctness} (see Figure~\ref{fig: sp_corr} for a representative example).
To quantify this effect, we use Gemini-3-Pro\footnote{\url{https://storage.googleapis.com/deepmind-media/Model-Cards/Gemini-3-Pro-Model-Card.pdf}} to assess whether the generated rationales are logically consistent with the corresponding golden judgments\footnote{The evaluation protocol and implementation details are provided in Section~\ref{sec:benchmark}.}.
The results show a substantial disparity between the two GenRMs: RRM-32B exhibits high spurious rates across benchmarks (59.0\% on HelpSteer3, 36.7\% on RewardBench2, and 62.4\% on PPE-Preference), whereas Qwen3-14B maintains markedly lower rates (40.0\%, 20.1\%, and 36.9\%, respectively). These findings suggest that RRM-32B frequently bases its decisions on unreliable heuristics, which in turn undermines its suitability for RLHF.

In summary, our analysis reveals a systematic disconnect between \emph{label accuracy} and \emph{judgment soundness} in GenRMs. We posit that this misalignment contributes to RLHF instability: during optimization, the policy can \textbf{amplify and exploit spurious cues} that are rewarded by the GenRM, instead of improving the underlying response quality. Motivated by this observation, Section~\ref{sec:benchmark} introduces a rationale-aware benchmark that explicitly measures the gap between superficial correctness and valid reasoning.

\section{Rationale-Aware GenRM Benchmarking}
\label{sec:benchmark}

Motivated by the phenomenon of Spurious Correctness identified in Section~\ref{sec:spurious_correctness}, we introduce a rationale-aware benchmarking framework designed to rigorously evaluate GenRMs. Moving beyond outcome-centric metrics, our approach verifies the logical consistency between the GenRMs' generated judgment and the ground truth. To support this fine-grained verification, we begin by defining the problem setting and key components of our framework.

\subsection{Formulation}

\textbf{Preference Dataset.}
Let $\mathcal{D} = \{(x, y_a, y_b, l)\}$ denote a preference dataset, where $x$ represents the prompt, $(y_a, y_b)$ is a pair of candidate responses, and $l \in \{a, b\}$ is the ground-truth label indicating the preferred response.

\textbf{Generative Reward Model.}
We denote the GenRM as $\mathcal{R}$. Given the prompt $x$ and candidate responses $(y_a, y_b)$, the model evaluates the pair by generating a judgment $\mathbf{o}$:
\begin{equation}
    \mathbf{o} \sim \mathcal{R}(\cdot \mid x, y_a, y_b)
\end{equation}
The output $\mathbf{o}$ contains the natural language analysis of the response quality. We then apply a deterministic parsing function $f(\cdot)$ to extract the discrete model verdict $\hat{l} = f(\mathbf{o})$ (where $\hat{l} \in \{a, b\}$) from the judgment, representing the GenRM's predicted preference.

\textbf{Meta-Reward Model (MetaRM).}
To supervise the rationale quality, we introduce a MetaRM $\mathcal{M}$. Let $\mathbf{o}^*$ denote the reference rationale from a golden judge. The MetaRM evaluates the input tuple $(x, y_a, y_b, \mathbf{o}^*, \mathbf{o})$ and generates an assessment, which is parsed into a binary alignment decision $v_{\text{meta}} \in \{0, 1\}$:
\begin{equation}
    v_{\text{meta}} \leftarrow \mathcal{M}(x, y_a, y_b, \mathbf{o}^*, \mathbf{o})
    \label{eq_metarm}
\end{equation}
Here, $v_{\text{meta}}=1$ signifies that the model's judgment $\mathbf{o}$ accurately captures the core reasoning of the reference $\mathbf{o}^*$, while $v_{\text{meta}}=0$ indicates a misalignment.

\subsection{The Meta-Judging Pipeline}
\label{sec:metarm_impl}

We implement the MetaRM using a three-stage Chain-of-Thought verification process (prompt details in Appendix~\ref{app:prompt_meta}). 
First, the model analyzes the golden rationale $\mathbf{o}^*$ to extract \textit{Key Discriminators}—the specific causal factors (e.g., factual errors or safety violations) that necessitate the preference label, isolating valid logic from boilerplate text. 
Second, it performs \textit{Rationale Coverage Verification} to determine if the GenRM's rationale $\mathbf{o}$ explicitly identifies these discriminators. This step strictly penalizes \textit{spurious correctness}; for instance, if $\mathbf{o}$ cites superficial stylistic issues while $\mathbf{o}^*$ points to a calculation error, the alignment is rejected. 
Finally, the MetaRM outputs a binary verdict $v_{\text{meta}} \in \{0,1\}$, where $v_{\text{meta}}=1$ signifies that the GenRM's reasoning is logically consistent with the golden reference, ensuring the model is rewarded for valid causal analysis.

\subsection{Benchmark Construction \& Metrics}

We augment a rationale-aware benchmark derived from {HelpSteer3}, {RewardBench2}, and {PPE-Preference}. Addressing the lack of unified critiques in these datasets, we employ {Gemini-3-Pro}~\citep{google2025gemini} to generate golden reference rationales $\mathbf{o}^*$. We adopt a two-fold strategy: (1) for label-only datasets ({RewardBench2}, {PPE-Preference}), the model generates reasoning conditioned on the ground-truth label $l$; (2) for {HelpSteer3}, it acts as a meta-reviewer to aggregate diverse human judgments into a coherent reference. The resulting benchmark consists of pairwise samples $(x, y_a, y_b, \mathbf{o}^*)$, with detailed procedures in Appendix~\ref{app:data_construction}. We employ Gemini-3-Pro as the MetaRM $\mathcal{M}$ for all evaluations to ensure high-fidelity rationale alignment detection. Its reliability is empirically validated against human annotators in Appendix~\ref{app:human}.

To quantify the gap between superficial preference prediction and genuine logical alignment, we define three key metrics evaluated on our rationale-aware benchmark. For a given GenRM $\mathcal{R}$ and a MetaRM $\mathcal{M}$, we denote $N$ as the total number of samples in the benchmark. 

\begin{enumerate}
    \item \textbf{Label Accuracy (L-Acc):} This is the standard metric used in existing reward model leaderboards, measuring the model's ability to predict the correct preference label.
    \begin{equation}
        \text{L-Acc} = \frac{1}{N} \sum_{i=1}^{N} \mathbb{I}(\hat{l}_i = l_i)
    \end{equation}

    \item \textbf{Spurious Correctness (S-Corr):} This is our core diagnostic metric. It measures the proportion of samples where the GenRM arrives at the correct verdict $\hat{l}=l$ but fails the MetaRM's logical verification ($v_{\text{meta}}=0$).
    \begin{equation}
        \text{S-Corr} = \frac{\sum_{i=1}^{N} \mathbb{I}(\hat{l}_i = l_i \land v_{\text{meta}, i} = 0)}{\sum_{i=1}^{N} \mathbb{I}(\hat{l}_i = l_i)}
    \end{equation}

    \item \textbf{Fidelity Score(F-Score):} This metric represents the most stringent evaluation, requiring the model to be correct in both its final decision and its underlying rationale.
    \begin{equation}
        \text{F-Score} = \frac{1}{N} \sum_{i=1}^{N} \mathbb{I}(\hat{l}_i = l_i \land v_{\text{meta}, i} = 1)
    \end{equation}
\end{enumerate}

\begin{table*}[ht]
    \centering
    \caption{Main results of diverse GenRMs on our Rational-Aware Benchmark. }
    \label{tab:model_comparison}
    \resizebox{\textwidth}{!}{
        \begin{tabular}{l ccc ccc ccc}
            \toprule
            \multirow{2}{*}{\textbf{Model}} & \multicolumn{3}{c}{\textbf{HelpSteer 3}} & \multicolumn{3}{c}{\textbf{RewardBench 2}} & \multicolumn{3}{c}{\textbf{PPE-Preference}} \\
            \cmidrule(lr){2-4} \cmidrule(lr){5-7} \cmidrule(lr){8-10}
             & L-Acc$\uparrow$ & S-Corr$\downarrow$ & F-Score$\uparrow$ & L-Acc $\uparrow$ & S-Corr$\downarrow$ & F-Score$\uparrow$ & L-Acc $\uparrow$ & S-Corr$\downarrow$ & F-Score$\uparrow$ \\
            \midrule
            
            \rowcolor{gray!15} 
            \multicolumn{10}{c}{\textbf{LLM-as-Judge}} \\
            \midrule
            
            DeepSeek-V3.2-chat         & 75.9 & 24.9 & 57.0 & 90.2 & 13.9 & 77.7 & 66.4 & 24.0 & 50.5 \\
            DeepSeek-V3.2-thinking     & 77.5 & 29.5 & 54.0 & 91.9 & 18.9 & 74.5 & 66.8 & 26.2 & 49.2 \\
            Gemini-2.5-Pro             & 78.4 & 12.4 & 68.6 & 90.1 & 4.8  & 85.7 & 69.8 & 8.3  & 64.0 \\
            Gemini-3-Pro               & 78.1 & 5.6  & 73.7 & 92.0 & 1.7  & 90.4 & 62.7 & 1.6  & 61.6 \\
            GPT-5-chat                 & 77.5 & 20.5 & 61.5 & 93.4 & 10.3 & 83.7 & 66.6 & 19.6 & 53.5 \\
            GPT-5-thinking             & 76.0 & 11.7 & 67.1 & 92.6 & 3.4  & 89.4 & 64.7 & 6.9  & 58.8 \\
            Claude-Sonnet-4.5          & 78.4 & 15.7 & 66.0 & 91.6 & 9.4  & 83.0 & 69.2 & 14.5 & 59.1 \\
            Claude-Sonnet-4.5-thinking & 79.9 & 12.5 & 69.9 & 93.1 & 5.1  & 88.3 & 68.7 & 8.7  & 62.7 \\

            \midrule
            GPT-OSS-120B           & 76.5 & 37.9 & 47.4 & 90.6 & 16.2 & 75.9 & 64.5 & 40.7 & 38.2 \\
            Qwen3-4B-Instruct-2507     & 72.8 & 44.8 & 40.2 & 85.3 & 27.4 & 61.9 & 60.4 & 44.8 & 33.3 \\
            Qwen3-4B-Thinking-2507     & 72.5 & 34.9 & 47.2 & 87.8 & 19.3 & 70.9 & 61.5 & 33.7 & 40.8 \\
            \midrule
            \rowcolor{gray!15} 
            \multicolumn{10}{c}{\textbf{Specialized Generative Reward Models}} \\
            \midrule
            
            RRM-32B   & 74.7 & 59.0 & 30.6 & 88.5 & 36.7 & 56.0 & 65.1 & 62.4 & 24.5 \\
            RM-R1-DS-32B   & 73.2 & 50.7 & 36.1 & 84.2 & 25.0 & 63.2 & 64.9 & 46.5 & 34.7  \\
            RM-R1-Qwen-32B & 75.1 & 32.9 & 50.4 & 87.1 & 18.3 & 71.2 & 65.6 & 30.0 & 45.9  \\
            
            \midrule
            \rowcolor{gray!15} 
            \multicolumn{10}{c}{\textbf{Our Methods}} \\
            \midrule
            Qwen3-8B        & 72.9 & 45.8 & 39.5 & 82.7 & 24.6 & 62.4 & 62.2 & 45.2 & 34.1 \\
            GenRM-RLVR-8B   & 72.9 & 44.6 & 40.4 & 89.2 & 25.9 & 66.1 & 63.7 & 47.4 & 33.5 \\
            \textbf{GenRM-R-Align-8B}  & 73.1 & 34.6 & 47.8 & 89.8 & 21.7 & 70.3 & 63.6 & 34.9 & 41.4 \\
            \midrule
            Qwen3-14B     & 74.1 & 40.0 & 44.5 & 87.9 & 20.1 & 70.2 & 65.2 & 36.9 & 41.1 \\
            GenRM-RLVR-14B  & 75.5 & 46.9 & 40.1 & 88.1 & 29.0 & 62.5 & 65.7 & 45.7 & 35.6 \\
            \textbf{GenRM-R-Align-14B} & 76.3 & 29.2 & 54.0 & 92.0 & 14.6 & 78.5 & 65.7 & 26.9 & 48.1 \\

            \bottomrule
        \end{tabular}%
    }
\end{table*}

\subsection{Quantifying Misalignment in GenRM}

We evaluate various GenRMs using the proposed metrics. Our analysis reveals several critical insights into the current landscape of reward modeling:

\textbf{Prevalence of Spurious Correctness.} Our analysis reveals that S-Corr is a prevalent issue across the current landscape of reward modeling, affecting both powerful proprietary models (e.g., GPT-5~\citep{openai2025gpt5systemcard}, DeepSeek-V3.2~\citep{liu2025deepseek}) and open-weights models of varying scales (e.g., Qwen3-8B, Qwen3-14B, GPT-OSS-120B~\citep{agarwal2025gpt}). However, a distinct trend emerges: as model capability increases, the reliance on spurious heuristics diminishes. For instance, larger or more advanced models consistently exhibit lower S-Corr compared to their smaller counterparts. Furthermore, the mode of reasoning plays a critical role; models with Chain-of-Thoughts (e.g., Qwen3-4B-Thinking-2507, GPT-5-thinking, Claude-Sonnet-4.5-thinking~\cite{anthropic2025claude45}) demonstrate a marked reduction in S-Corr compared to their non-thinking variants (e.g., Qwen3-4B-Instruct-2507, GPT-5-chat,  Claude-Sonnet-4.5). This reduction in spurious correctness directly translates to higher F-Score, indicating that stronger reasoning capabilities enable models to align not just with the final preference, but with the correct underlying logic.

\textbf{The Fragility of Standard Benchmarks.} In sharp contrast to the stratification revealed by S-Corr, standard Label Accuracy (L-Acc) exhibits significant saturation and fails to effectively discriminate between models. As shown in Table~\ref{tab:model_comparison}, L-Acc scores are often tightly clustered regardless of model capacity: most models achieve scores in the 70\%+ range on HelpSteer3 and hover around 90\% on RewardBench2, while scores on PPE-Preference largely stagnate in the 60\%+ range for open-source models. This suggests that traditional binary accuracy has become an insufficient proxy for reward model quality. F-Score, by enforcing a strict alignment between the generated rationale and the golden rationale, breaks this ceiling and offers a far more granular metric for model differentiation. We further validate the practical superiority of F-Score by correlating it with downstream RLHF performance in Section~\ref{sec:rlhf_exp}.

\section{Aligning GenRMs via Meta-Judging}
\label{sec:R-align}

To address the misalignment revealed in Section~\ref{sec:benchmark}, we propose a Rationale-Centric Alignment objective that enforces consistency in both the final verdict and the judgment rationale.

\subsection{Rationale-Aware Reward}
\textbf{Reward Function Formulation.} 
We optimize the GenRM $\mathcal{R}$ using reinforcement learning. Consider a standard Outcome-Supervised Reward (Baseline), which depends solely on the verdict correctness:
\begin{equation}
    \label{eq:outcome}
    R_{\text{outcome}} = 
    \begin{cases} 
    1 & \text{if } \hat{l} = l \\
    0 & \text{otherwise} 
    \end{cases}
\end{equation}
This reward ignores the quality of the generated judgment $\mathbf{o}$. In contrast, our proposed Rationale-Aware Reward incorporates the MetaRM's assessment $v_{\text{meta}}$ to verify the judgment alignment. The reward is positive only when \textit{both} the verdict is correct and the judgment is verified by the MetaRM:
\begin{equation}
    \label{eq:overall}
    R_{\text{overall}} = 
    \begin{cases} 
    1 & \text{if } \hat{l} = l \land v_{\text{meta}} = 1 \\
    0 & \text{otherwise} 
    \end{cases}
\end{equation}
Here, $v_{\text{meta}} \in \{0, 1\}$ is the alignment decision from the MetaRM (as defined in Eq. ~\ref{eq_metarm}). This strict reward mechanism penalizes instances where the model guesses the correct label with a flawed analysis (i.e., $\hat{l}=l$ but $v_{\text{meta}}=0$), effectively aligning the GenRM's judgment process with the reference standard.

\begin{figure*}[t]
    \centering
    
    \includegraphics[width=\textwidth]{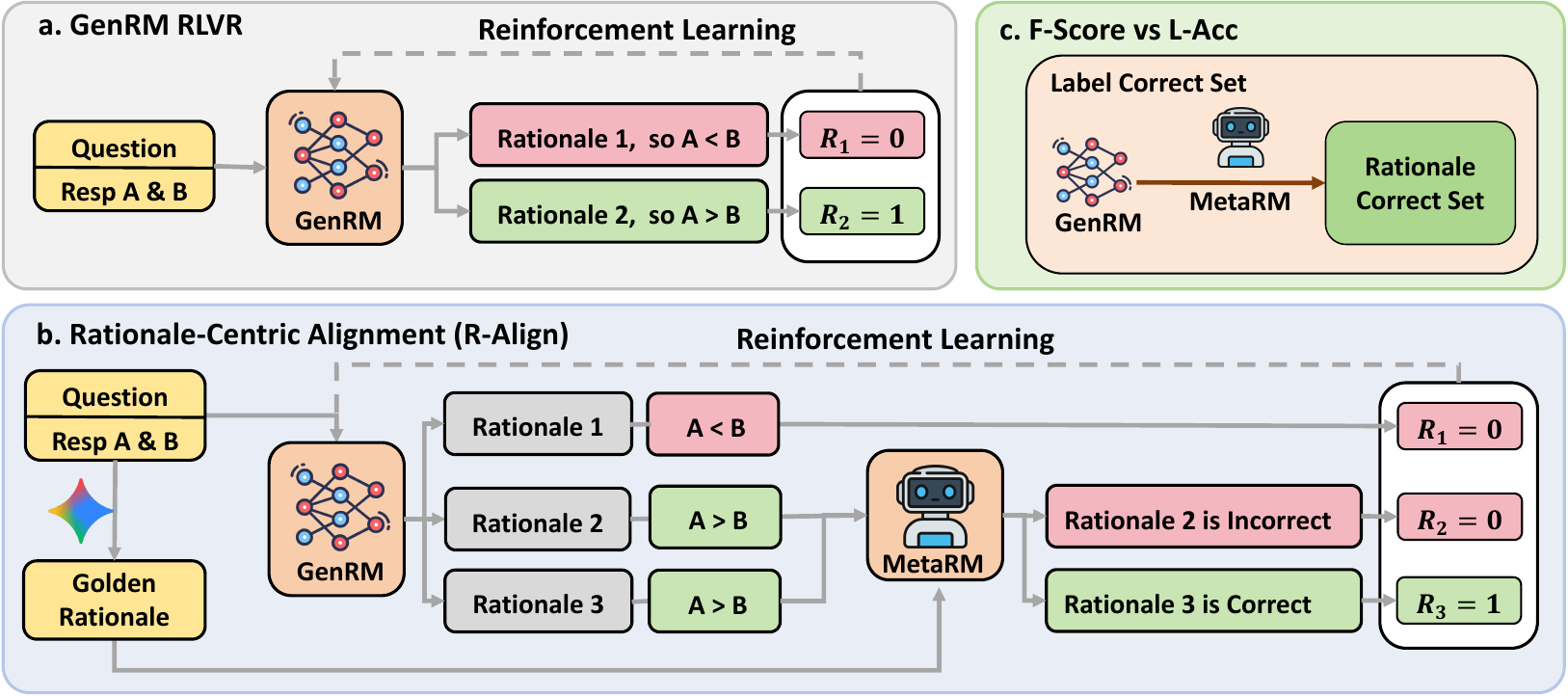}
    \captionsetup{singlelinecheck=false, justification=raggedright}
    \caption{Overview of the MetaRM framework. (a) GenRM RLVR (Baseline): The model is optimized solely on outcome correctness, receiving rewards ($R=1$) for accurate preference labels regardless of the reasoning quality. (b) Rationale-Centric Alignment (Ours): Incorporates a MetaRM to enforce process supervision; rewards are granted only when both the label is correct and the rationale is logically consistent with the golden reference, effectively penalizing spurious correctness. (c) F-Score vs. L-Acc: Visualizes F-Score as a strict subset of L-Acc, filtering out spurious correctness where the model is right for the wrong reasons.}
    \label{fig:pipeline}
\end{figure*}

\begin{figure*}[t]
    \centering
    \begin{subfigure}{0.48\textwidth}
        \centering
        \includegraphics[width=\textwidth]{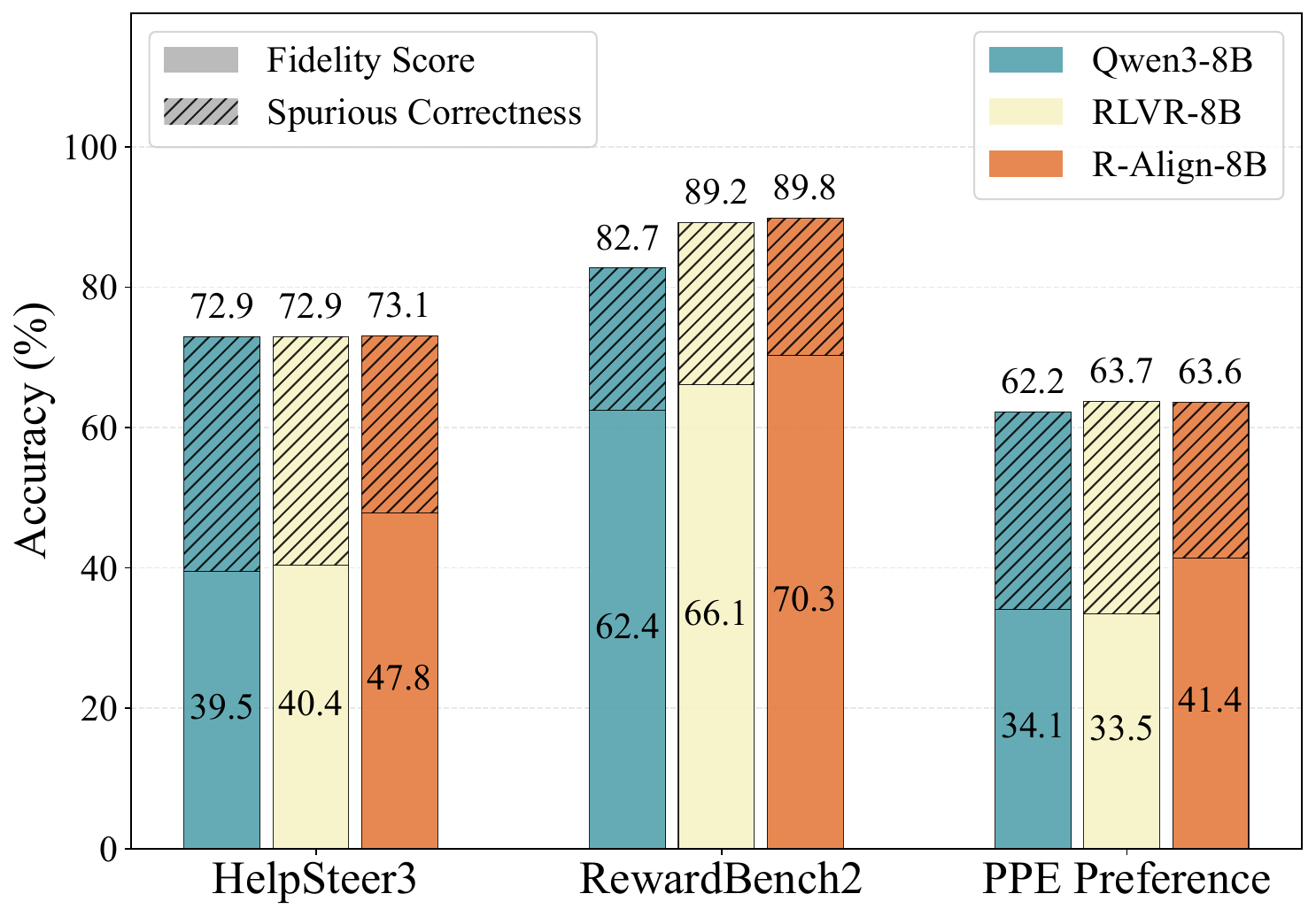}
        \caption{Benchmark results for 8B models.}
        \label{fig:eval_8b}
    \end{subfigure}
    \hfill 
    \begin{subfigure}{0.48\textwidth}
        \centering
        \includegraphics[width=\textwidth]{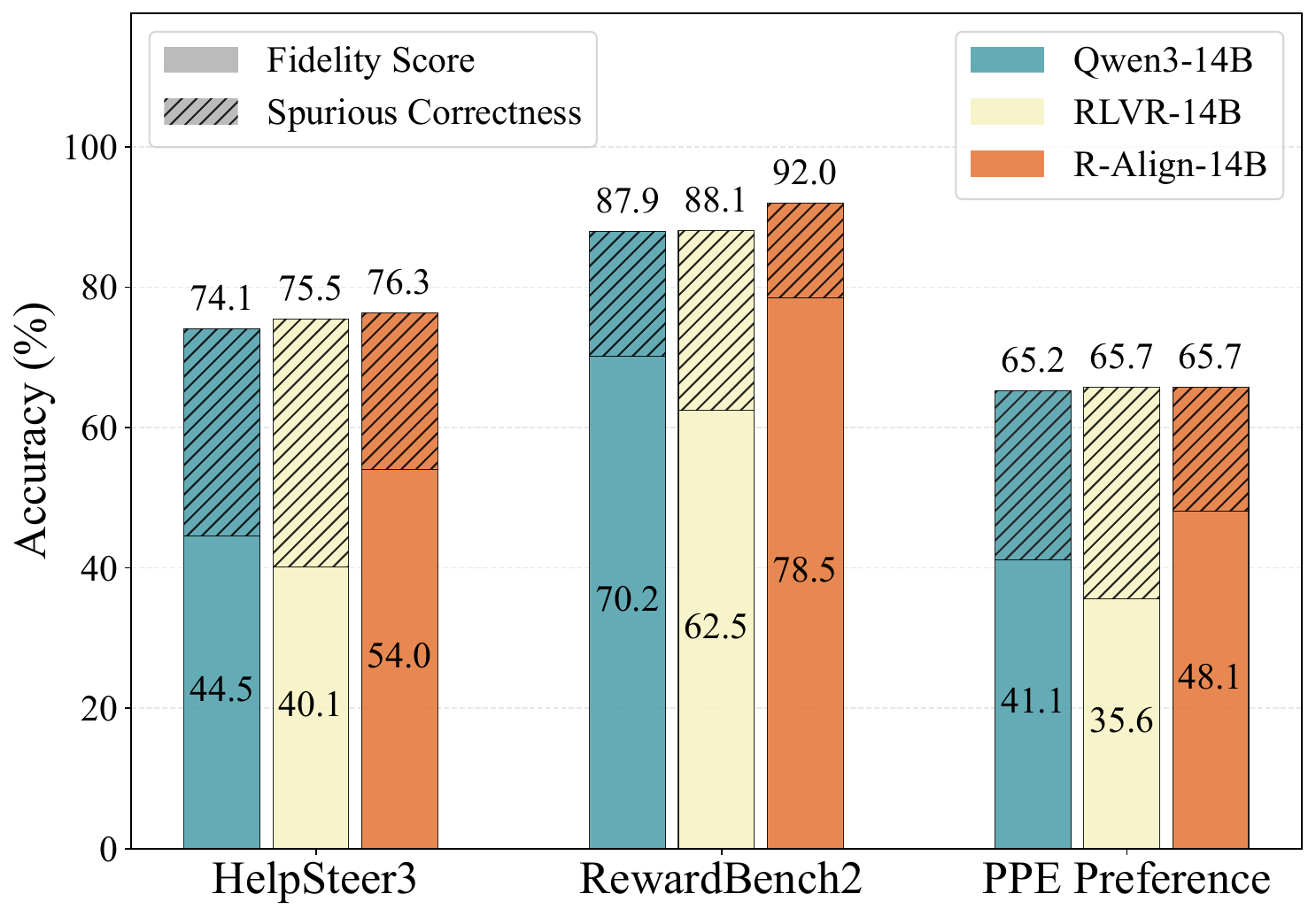}
        \caption{Benchmark results for 14B models.}
        \label{fig:eval_14b}
    \end{subfigure}
    \captionsetup{singlelinecheck=false, justification=raggedright}
    \caption{Benchmark results on HelpSteer3, RewardBench2, and PPE-Preference. The numerical labels on top of the bars denote the Label Accuracy. The solid bars represent the F-Score, while the hatched areas indicate the proportion of Spurious Correctness.}
    \label{fig:combined_eval}
\end{figure*}

\subsection{Training Implementation} 
\textbf{Models.} 
We employ the {Qwen3-8B} and {Qwen3-14B} as our initialization checkpoints for GenRM training. These models serve as the policy $\mathcal{R}$ in our RL experiments.

\textbf{Training Data.} 
Following RM-R1~\cite{chen2025rm}, we utilize a cleaned subset of Skywork Reward Preference 80K~\cite{liu2024skywork}, 8K samples from Code-Preference-Pairs, and the complete Math-DPO-10K dataset~\cite{lai2024step}. Additionally, we incorporate the HelpSteer3~\citep{wang2025helpsteer3} training dataset, which contains high-quality human preference data.
To enable process supervision, we augment all training samples with golden judgments $\mathbf{o}^*$. These references are generated by {Gemini-3-Pro} following our benchmark construction in Section~\ref{sec:benchmark}: generating rationales for label-only datasets and aggregating multi-annotator reviews for {HelpSteer3}, resulting in tuples $(x, y_a, y_b, l, \mathbf{o}^*)$.

\textbf{Baselines \& Training.} 
We compare our proposed method against a strong RLVR baseline using the PPO algorithm. 
(1) \textbf{GenRM-RLVR:} The baseline GenRM is trained solely with $R_{\text{outcome}}$, receiving a positive reward whenever the predicted verdict matches the ground truth label, regardless of the reasoning quality.
(2) \textbf{GenRM-R-Align:} The GenRM is trained with the MetaRM-based reward $R_{\text{overall}}$, which penalizes spurious correctness by requiring both verdict accuracy and reasoning alignment.
All hyperparameters are kept consistent across runs to isolate the impact of process supervision.

\begin{wraptable}{r}{0.5\textwidth}
    \centering
    \captionsetup{font=small}
    \caption{Meta-judging agreement (F1) of open-weights models with Gemini-3-Pro}
    \label{tab:consistency}
    
    \resizebox{\linewidth}{!}{
        \begin{tabular}{lccc}
            \toprule
            MetaRM & RewardBench2 & Helpsteer3 & PPE-Preference \\
            \midrule
            Qwen3-8B         & 89.35 & 76.43 & 77.18 \\
            Qwen3-14B        & 89.49 & 81.22 & 80.76 \\
            Qwen3-32B        & 86.47 & 78.35 & 77.93 \\
            GPT-OSS-120B     & 90.68 & 83.53 & 84.33 \\
            \bottomrule
        \end{tabular}
    }
\end{wraptable}
\paragraph{MetaRM Selection.} 

While we employ Gemini-3-Pro to provide meta-judging for benchmarking(whose reliability is validated against human annotators in Appendix~\ref{app:human}), deploying such a proprietary model for the high-frequency queries inherent to RL training is prohibitively expensive. 
To identify a scalable open-weight alternative, we evaluate various models against Gemini-3-Pro's verdicts. As shown in Table~\ref{tab:consistency}, GPT-OSS-120B exhibits the highest alignment (e.g., 90.7 F1 on RewardBench2), significantly outperforming the Qwen3 series. Consequently, we select GPT-OSS-120B as the training MetaRM to ensure scalable, high-fidelity verification.

\subsection{Benchmark Performance}
Table~\ref{tab:model_comparison} and Figure~\ref{fig:combined_eval} present the evaluation results, comparing base models, standard outcome supervision (GenRM-RLVR), and our proposed rationale-centric alignment (GenRM-R-Align) across {HelpSteer3}, {RewardBench2}, and {PPE-Preference}.

\textbf{The Uncoupling of Label Accuracy and Reasoning Quality.}
While standard RLVR improves L-Acc, it fails to address rationale misalignment (visually represented by the hatched areas in Figure~\ref{fig:combined_eval}). For instance, on {HelpSteer3}, applying RLVR to Qwen3-14B boosts L-Acc (74.1\% $\to$ 75.5\%) but simultaneously spikes S-Corr (40.0\% $\to$ 46.9\%), causing F-Score to drop from 44.5\% to 40.1\%. A similar trade-off appears on {PPE-Preference} with Qwen3-8B, where L-Acc gains (+1.5\%) come at the cost of reasoning quality. This confirms that optimizing solely for outcome correctness incentivizes superficial heuristics over robust judgment logic.

\textbf{Effectiveness of R-Align.}
R-Align significantly mitigates this pathology, consistently achieving the lowest S-Corr and highest F-Score across all settings. On {RewardBench2}, GenRM-R-Align-8B attains an F-Score of 70.3\%, surpassing the larger GenRM-RLVR-14B baseline by 4.2\% while reducing S-Corr by the same margin. Notably, on {HelpSteer3}, our method boosts Qwen3-14B's F-Score to 54.0\% (+13.9\% over RLVR), demonstrating that penalizing spurious correctness effectively bridges the gap between preference prediction and logical entailment.

\section{Downstream RLHF Performance}
In this section, we evaluate the performance of the GenRMs trained in Section \ref{sec:R-align} by utilizing them to supervise the downstream RLHF training of policy models.

\subsection{Downstream RLHF Setup.}
\label{sec:exp_setup}

We conduct our downstream RLHF experiments using the Arena-Human-Preference dataset~\citep{chiang2024chatbot} as the source of training prompts. We initialize the policy with Qwen3-8B. To evaluate the performance of the aligned actors, we employ a comprehensive suite of benchmarks covering diverse domains: AIME24~\cite{aime24} and AIME25~\cite{aime25} for mathematics; GPQA-diamond~\cite{rein2024gpqa} for expert-level reasoning; LiveCodeBench~\cite{jain2024livecodebench} for coding capabilities; and MultiChallenge~\cite{deshpande2025multichallenge}, Arena-Hard-v2~\cite{li2024crowdsourced}, Wildbench~\cite{lin2024wildbench}, and IFBench~\cite{pyatkin2025generalizing} for general instruction following and conversational ability. Detailed implementation settings, including the training algorithm, reward formulation, and length control mechanisms, are provided in Appendix~\ref{app:rlhf_details}.

\begin{table*}[ht]
    \centering
    \captionsetup{font=small}
    \caption{Comparative analysis of performance gains for Qwen3-8B enhanced by different GenRMs. \textbf{Abbreviations:} LCB: LiveCodeBench, WB: WildBench, AH2: Arena-Hard-V2, MC: MultiChallenge.}
    \label{tab:downstream}

    \small 
    \setlength{\tabcolsep}{3.5pt} 
    
    \begin{NiceTabular}{l |ccc cc ccc |c}
        \toprule
        \multirow{2}{*}{\textbf{Reward Model}} & \multicolumn{3}{c}{\textbf{STEM}} & \multicolumn{1}{c}{\textbf{Code}} & \multicolumn{1}{c}{\textbf{IF}} & \multicolumn{3}{c}{\textbf{General}} & \multirow{2}{*}{\textbf{AVG}}\\
        \cmidrule(lr){2-4} \cmidrule(lr){5-5} \cmidrule(lr){6-6} \cmidrule(lr){7-9}

         & AIME24 & AIME25 & GPQA & LCB & IFBench & WB & AH2 & MC \\
        \midrule
        Qwen3-8B             & 77.6 & 67.6 & 60.9 & 50.8 & 32.0 & 72.8 & 26.5 & 42.9 & 53.9 \\
        \midrule
        + Qwen3-8B-as-GenRM             & 77.7 & 64.8 & 59.7 & 48.8 & 35.0 & 79.4 & 46.6 & 52.8 & 58.1 \\
        + GenRM-RLVR-8B        & 73.7 & 58.3 & 58.7 & 47.4 & 30.3 & 84.3 & 53.1 & 55.0 & 57.6 \\
        + \textbf{GenRM-R-Align-8B} & 75.4 & 64.2 & 59.0 & 51.5 & 26.9 & 89.2 & 59.5 & 51.7 & \textbf{59.7} \\
        \midrule
        + Qwen3-14B-as-GenRM              & 76.1 & 64.1 & 59.5 & 50.5 & 29.6 & 83.7 & 51.0 & 58.6 & 59.1 \\
        + GenRM-RLVR-14B         & 75.5 & 64.8 & 57.1 & 50.1 & 31.6 & 88.1 & 55.9 & 51.7 & 59.4 \\
        + \textbf{GenRM-R-Align-14B}  & 76.5 & 67.2 & 60.3 & 49.4 & 31.6 & 92.6 & 60.2 & 55.7 & \textbf{61.7} \\
        \bottomrule
    \end{NiceTabular}
\end{table*}
\subsection{Policy Performance}
\label{sec:rlhf_exp}

\paragraph{Overall Performance and Trade-offs.}

First, our RLHF pipeline is effective overall. Even when we use the base model itself as the reward model ({Qwen3-8B-as-GenRM}), the resulting policy achieves a clear improvement in average performance. Since Qwen3-8B already exhibits strong initial capability on STEM and code benchmarks, the largest RLHF gains primarily appear in general-purpose behavior, reflected by consistent improvements on WildBench, Arena-Hard-v2, and MultiChallenge. However, these improvements often coincide with degradations in specialized STEM/code reasoning, illustrating the well-known alignment–capability trade-off (the “alignment tax”)~\citep{ouyang2022training, lin2024mitigatingalignmenttaxrlhf, chaudhari2024rlhfdecipheredcriticalanalysis}.

\paragraph{Effectiveness of R-Align.}
Comparing the training paradigms, our proposed GenRM-R-Align consistently outperforms the standard outcome-centric baseline ({GenRM-RLVR}) across nearly all evaluated domains. The advantage of incorporating rationale supervision manifests in two critical aspects:

\begin{itemize}
    \item \textbf{Mitigating the Alignment Tax in Reasoning:} 
    Standard outcome supervision often leads to severe performance regression in STEM-related tasks. We observe that the policy trained with {GenRM-RLVR-8B} suffers significant drops on benchmarks like LiveCodeBench (47.4) and AIME25 (58.3). 
    In sharp contrast, R-Align effectively preserves the model's reasoning capabilities, recovering performance to 75.4 and 64.2 respectively.
    
    \item \textbf{Boosting General Capabilities:} 
    Beyond preserving reasoning, our method drives substantial improvements in the general domain. On benchmarks such as WildBench and Arena-Hard-v2, policies trained with GenRM-R-Align significantly surpass the {GenRM-RLVR} baseline (e.g., achieving 89.2 vs. 84.3 on WildBench). This indicates that by verifying the reasoning process, R-Align provides a more robust and generalized preference signal than outcome supervision alone.
\end{itemize}

\subsection{Correlation with Downstream Performance}

\begin{wraptable}{r}{0.6\textwidth}
    \centering
    \caption{Pearson correlation coefficient between RLHF performance and F-Score/L-Acc across benchmarks.}
    \label{tab:corr}
    
    \resizebox{\linewidth}{!}{
        \begin{tabular}{lccc}
            \toprule
            Metric & HelpSteer3 & RewardBench2 & PPE-Preference \\
            \midrule
            L-Acc & 0.366 & 0.382 & 0.220 \\
            F-Score & \textbf{0.947} & \textbf{0.924} & \textbf{0.963} \\
            \bottomrule
        \end{tabular}
    }
    \vspace{-10pt}
\end{wraptable}

\begin{figure}[htbp]
    \centering
    \includegraphics[width=0.8\textwidth]{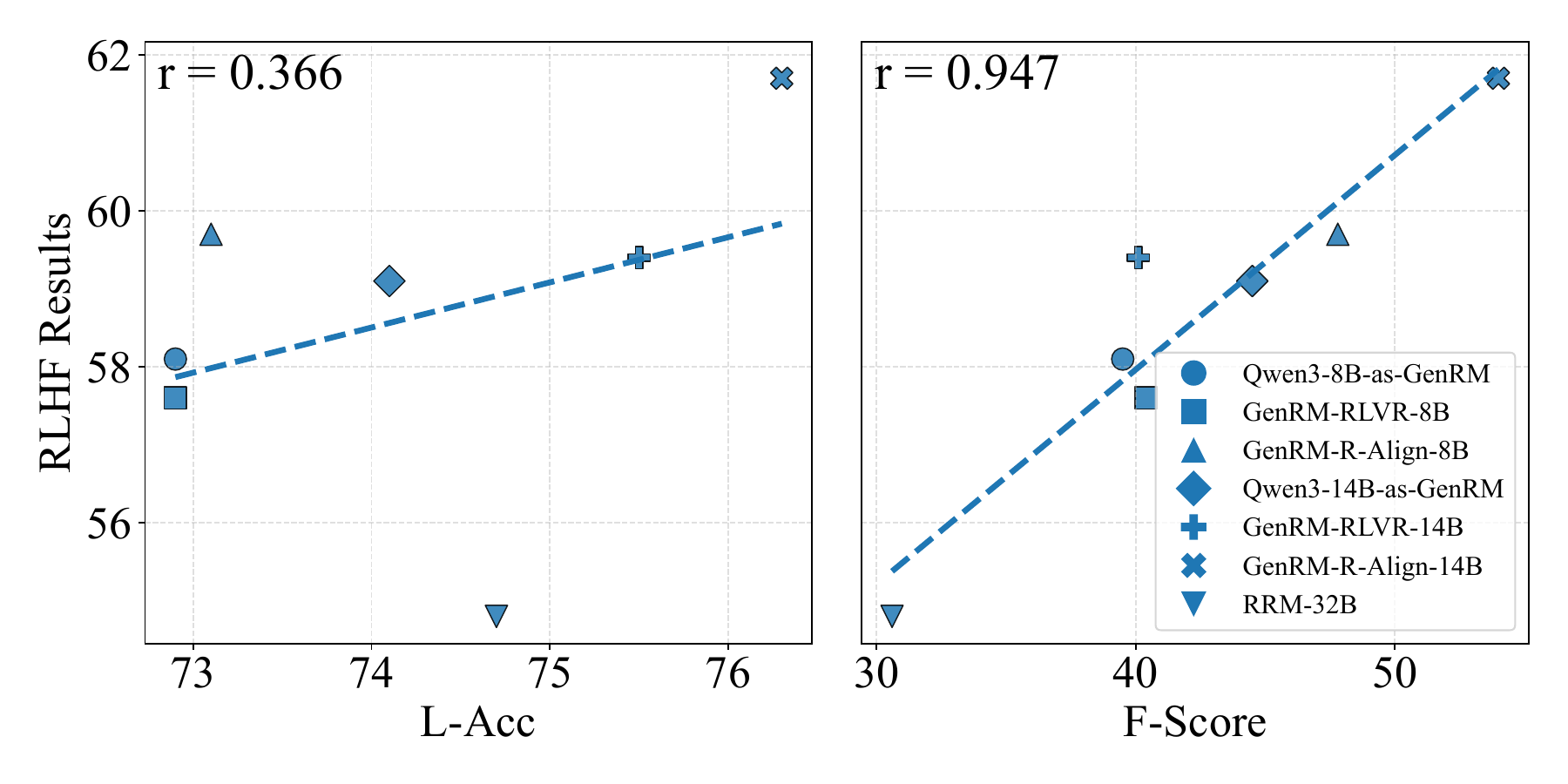}
    \captionsetup{singlelinecheck=false, justification=raggedright}
    \caption{
        Correlation analysis between benchmark metrics on \textbf{HelpSteer3} and downstream RLHF performance. 
        \textbf{Left:} Label Accuracy (L-Acc). 
        \textbf{Right:} Fidelity Score (F-Score).
    }
    \label{fig:corr_helpsteer}
\end{figure}

We compute the Pearson correlation coefficient between the average downstream RLHF performance of policies trained by the evaluated GenRMs and their benchmark scores (L-Acc and F-Score). The analysis covers all variants, including Qwen3 and RRM baselines along with our proposed models.

The results, presented in Table~\ref{tab:corr}, reveal a consistent trend: standard Label Accuracy (L-Acc) exhibits weak predictive power for downstream performance, with consistently low correlation coefficients (e.g., 0.366 on HelpSteer3, 0.382 on RewardBench2, and 0.220 on PPE-Preference). This indicates that outcome correctness alone is an insufficient proxy for reward quality, as it ignores the noisy supervision signals arising from spurious correlations. In contrast, F-Score demonstrates significantly higher correlations across the board, validating that rationale alignment is a more universal and robust predictor of effective policy guidance. 
This suggests that F-Score successfully bridges the disconnect often observed between offline proxy metrics and actual online RLHF outcomes.

Figure~\ref{fig:corr_helpsteer} visualizes this disparity on HelpSteer3. The left panel shows that L-Acc saturates in the 73\%--76\% range with a weak correlation ($r=0.366$), failing to effectively discriminate between models. Conversely, the right panel reveals that F-Score maintains a strong linear relationship ($r=0.947$) with downstream performance, confirming that enforcing logical consistency effectively filters out spurious correctness and provides a cleaner learning signal.

\section{Related Work}

\paragraph{Generative Reward Models}
Traditional reward modeling primarily relies on scalar reward models trained under Bradley-Terry model assumption, which regress a scalar score to represent human preference \cite{ouyang2022training,liu2024skywork}. The emergence of LLM-as-a-Judge marked a paradigm shift, utilizing the inherent reasoning capabilities of LLMs to evaluate responses via prompting \cite{zheng2023judging, bai2022constitutional}. Recent studies generally utilize RLVR to optimize GenRMs for accurate preference label prediction \cite{chen2025rm, guo2025reward, jiao2025think, chen2025reasongrm}. 
However, they typically lack explicit supervision over rationales, relying heavily on outcome-centric signals while leaving validity and logical consistency of reasoning processes unchecked.

\paragraph{Evaluating Reward Models}

RewardBench~\cite{lambert2025rewardbench} emphasizes the assessment of reward models on preference differences that are \textit{subtle yet verifiable}. 
Concurrently, RM-bench~\cite{liu2024rm} scrutinizes the robustness of RMs against subtle content variations and stylistic biases. 
Regarding the predictive power of these benchmarks, PPE~\cite{frick2024evaluate} investigates the correlation between RM evaluation metrics and downstream performance via Direct Preference Optimization (DPO)~\cite{rafailov2023direct}. However, their analysis is limited to the static training data inherent to offline methods. 
Most recently, addressing issues of score saturation and data contamination, RewardBench2~\cite{malik2025rewardbench} introduces unseen human prompts and increased difficulty levels. 
Crucially, while they observe a strong correlation between benchmark scores and Best-of-N (BoN) performance, they report a discrepancy in RLHF settings, finding that current benchmark metrics often fail to predict the performance of the downstream policy.

\section{Conclusion}

This paper identifies Spurious Correctness as a critical pathology hindering the effectiveness of GenRMs in RLHF, revealing that standard Label Accuracy often masks the model's reliance on spurious correlations. To mitigate this, we introduce the Rationale-Aware benchmarking for evaluating logical consistency. And we propose the Rationale-Centric Alignment (R-Align) training framework, which utilizes Meta-Judging to enforce rationale-based alignment. Our results demonstrate that prioritizing rationale integrity over simple label accuracy effectively filters out spurious correlations, thereby enhancing the stability and performance of downstream policy models. Ultimately, this work underscores the necessity of shifting from outcome-centric to rationale-centric supervision for robust Large Language Model alignment. Future work may further explore how such rationale-centric frameworks can be applied to superalignment in domains where human labeling is scarce.

\newpage
\bibliography{main}

\begin{thebibliography}{43}
\providecommand{\natexlab}[1]{#1}
\providecommand{\url}[1]{\texttt{#1}}
\expandafter\ifx\csname urlstyle\endcsname\relax
  \providecommand{\doi}[1]{doi: #1}\else
  \providecommand{\doi}{doi: \begingroup \urlstyle{rm}\Url}\fi

\bibitem[Agarwal et~al.(2025)Agarwal, Ahmad, Ai, Altman, Applebaum, Arbus, Arora, Bai, Baker, Bao, et~al.]{agarwal2025gpt}
S.~Agarwal, L.~Ahmad, J.~Ai, S.~Altman, A.~Applebaum, E.~Arbus, R.~K. Arora, Y.~Bai, B.~Baker, H.~Bao, et~al.
\newblock gpt-oss-120b \& gpt-oss-20b model card.
\newblock \emph{arXiv preprint arXiv:2508.10925}, 2025.

\bibitem[{Anthropic}(2025)]{anthropic2025claude45}
{Anthropic}.
\newblock {Claude Sonnet 4.5 System Card}.
\newblock System card, Anthropic, Sept. 2025.
\newblock URL \url{https://assets.anthropic.com/m/12f214efcc2f457a/original/Claude-Sonnet-4-5-System-Card.pdf}.
\newblock Accessed: 2026-01-05.

\bibitem[Bai et~al.(2022)Bai, Kadavath, Kundu, Askell, Kernion, Jones, Chen, Goldie, Mirhoseini, McKinnon, et~al.]{bai2022constitutional}
Y.~Bai, S.~Kadavath, S.~Kundu, A.~Askell, J.~Kernion, A.~Jones, A.~Chen, A.~Goldie, A.~Mirhoseini, C.~McKinnon, et~al.
\newblock Constitutional ai: Harmlessness from ai feedback.
\newblock \emph{arXiv preprint arXiv:2212.08073}, 2022.

\bibitem[Chaudhari et~al.(2024)Chaudhari, Aggarwal, Murahari, Rajpurohit, Kalyan, Narasimhan, Deshpande, and da~Silva]{chaudhari2024rlhfdecipheredcriticalanalysis}
S.~Chaudhari, P.~Aggarwal, V.~Murahari, T.~Rajpurohit, A.~Kalyan, K.~Narasimhan, A.~Deshpande, and B.~C. da~Silva.
\newblock Rlhf deciphered: A critical analysis of reinforcement learning from human feedback for llms, 2024.
\newblock URL \url{https://arxiv.org/abs/2404.08555}.

\bibitem[Chen et~al.(2025{\natexlab{a}})Chen, Gao, Hu, Yu, Zhang, and Bao]{chen2025reasongrm}
B.~Chen, X.~Gao, C.~Hu, P.~Yu, H.~Zhang, and B.-K. Bao.
\newblock Reasongrm: Enhancing generative reward models through large reasoning models.
\newblock \emph{arXiv preprint arXiv:2506.16712}, 2025{\natexlab{a}}.

\bibitem[Chen et~al.(2025{\natexlab{b}})Chen, Li, Wang, Jin, Qian, Wang, Wang, Zhang, Zhang, Zhang, et~al.]{chen2025rm}
X.~Chen, G.~Li, Z.~Wang, B.~Jin, C.~Qian, Y.~Wang, H.~Wang, Y.~Zhang, D.~Zhang, T.~Zhang, et~al.
\newblock Rm-r1: Reward modeling as reasoning.
\newblock \emph{arXiv preprint arXiv:2505.02387}, 2025{\natexlab{b}}.

\bibitem[Chiang et~al.(2024)Chiang, Zheng, Sheng, Angelopoulos, Li, Li, Zhu, Zhang, Jordan, Gonzalez, et~al.]{chiang2024chatbot}
W.-L. Chiang, L.~Zheng, Y.~Sheng, A.~N. Angelopoulos, T.~Li, D.~Li, B.~Zhu, H.~Zhang, M.~Jordan, J.~E. Gonzalez, et~al.
\newblock Chatbot arena: An open platform for evaluating llms by human preference.
\newblock In \emph{Forty-first International Conference on Machine Learning}, 2024.

\bibitem[Deshpande et~al.(2025)Deshpande, Sirdeshmukh, Mols, Jin, Hernandez-Cardona, Lee, Kritz, Primack, Yue, and Xing]{deshpande2025multichallenge}
K.~Deshpande, V.~Sirdeshmukh, J.~B. Mols, L.~Jin, E.-Y. Hernandez-Cardona, D.~Lee, J.~Kritz, W.~E. Primack, S.~Yue, and C.~Xing.
\newblock Multichallenge: A realistic multi-turn conversation evaluation benchmark challenging to frontier llms.
\newblock In \emph{Findings of the Association for Computational Linguistics: ACL 2025}, pages 18632--18702, 2025.

\bibitem[Frick et~al.(2024)Frick, Li, Chen, Chiang, Angelopoulos, Jiao, Zhu, Gonzalez, and Stoica]{frick2024evaluate}
E.~Frick, T.~Li, C.~Chen, W.-L. Chiang, A.~N. Angelopoulos, J.~Jiao, B.~Zhu, J.~E. Gonzalez, and I.~Stoica.
\newblock How to evaluate reward models for rlhf.
\newblock \emph{arXiv preprint arXiv:2410.14872}, 2024.

\bibitem[Gao et~al.(2023)Gao, Schulman, and Hilton]{gao2023scaling}
L.~Gao, J.~Schulman, and J.~Hilton.
\newblock Scaling laws for reward model overoptimization.
\newblock In \emph{International Conference on Machine Learning}, pages 10835--10866. PMLR, 2023.

\bibitem[{Google DeepMind}(2025)]{google2025gemini}
{Google DeepMind}.
\newblock {Gemini: The Most Capable and General Model We've Ever Built}, 2025.
\newblock URL \url{https://deepmind.google/models/gemini/}.
\newblock Accessed: 2026-01-05.

\bibitem[Grattafiori et~al.(2024)Grattafiori, Dubey, Jauhri, Pandey, Kadian, Al-Dahle, Letman, Mathur, Schelten, Vaughan, Yang, Fan, Goyal, Hartshorn, Yang, et~al.]{grattafiori2024llama3}
A.~Grattafiori, A.~Dubey, A.~Jauhri, A.~Pandey, A.~Kadian, A.~Al-Dahle, A.~Letman, A.~Mathur, A.~Schelten, A.~Vaughan, A.~Yang, A.~Fan, A.~Goyal, A.~Hartshorn, A.~Yang, et~al.
\newblock The llama 3 herd of models.
\newblock \emph{arXiv preprint arXiv:2407.21783}, 2024.
\newblock URL \url{https://arxiv.org/abs/2407.21783}.

\bibitem[Guo et~al.(2025{\natexlab{a}})Guo, Yang, Zhang, Song, Wang, Zhu, Xu, Zhang, Ma, Bi, et~al.]{guo2025deepseek}
D.~Guo, D.~Yang, H.~Zhang, J.~Song, P.~Wang, Q.~Zhu, R.~Xu, R.~Zhang, S.~Ma, X.~Bi, et~al.
\newblock Deepseek-r1 incentivizes reasoning in llms through reinforcement learning.
\newblock \emph{Nature}, 645\penalty0 (8081):\penalty0 633--638, 2025{\natexlab{a}}.

\bibitem[Guo et~al.(2025{\natexlab{b}})Guo, Chi, Dong, Dong, Wu, Huang, and Wei]{guo2025reward}
J.~Guo, Z.~Chi, L.~Dong, Q.~Dong, X.~Wu, S.~Huang, and F.~Wei.
\newblock Reward reasoning model.
\newblock \emph{arXiv preprint arXiv:2505.14674}, 2025{\natexlab{b}}.

\bibitem[Huang et~al.(2026)Huang, Yao, Han, Wan, Guo, Lv, Zhou, Wang, Zhou, Sun, et~al.]{huang2026step3}
A.~Huang, C.~Yao, C.~Han, F.~Wan, H.~Guo, H.~Lv, H.~Zhou, J.~Wang, J.~Zhou, J.~Sun, et~al.
\newblock Step3-vl-10b technical report.
\newblock \emph{arXiv preprint arXiv:2601.09668}, 2026.

\bibitem[Jain et~al.(2024)Jain, Han, Gu, Li, Yan, Zhang, Wang, Solar-Lezama, Sen, and Stoica]{jain2024livecodebench}
N.~Jain, K.~Han, A.~Gu, W.-D. Li, F.~Yan, T.~Zhang, S.~Wang, A.~Solar-Lezama, K.~Sen, and I.~Stoica.
\newblock Livecodebench: Holistic and contamination free evaluation of large language models for code.
\newblock \emph{arXiv preprint arXiv:2403.07974}, 2024.

\bibitem[Jiao et~al.(2025)Jiao, Zeng, Vialard, Kuchaiev, Han, and Delalleau]{jiao2025think}
Y.~Jiao, J.~Zeng, J.~V. Vialard, O.~Kuchaiev, J.~Han, and O.~Delalleau.
\newblock Think twice: Branch-and-rethink reasoning reward model.
\newblock \emph{arXiv preprint arXiv:2510.23596}, 2025.

\bibitem[Lai et~al.(2024)Lai, Tian, Chen, Yang, Peng, and Jia]{lai2024step}
X.~Lai, Z.~Tian, Y.~Chen, S.~Yang, X.~Peng, and J.~Jia.
\newblock Step-dpo: Step-wise preference optimization for long-chain reasoning of llms.
\newblock \emph{arXiv preprint arXiv:2406.18629}, 2024.

\bibitem[Lambert et~al.(2025)Lambert, Pyatkin, Morrison, Miranda, Lin, Chandu, Dziri, Kumar, Zick, Choi, et~al.]{lambert2025rewardbench}
N.~Lambert, V.~Pyatkin, J.~Morrison, L.~J.~V. Miranda, B.~Y. Lin, K.~Chandu, N.~Dziri, S.~Kumar, T.~Zick, Y.~Choi, et~al.
\newblock Rewardbench: Evaluating reward models for language modeling.
\newblock In \emph{Findings of the Association for Computational Linguistics: NAACL 2025}, pages 1755--1797, 2025.

\bibitem[Li et~al.(2024)Li, Chiang, Frick, Dunlap, Wu, Zhu, Gonzalez, and Stoica]{li2024crowdsourced}
T.~Li, W.-L. Chiang, E.~Frick, L.~Dunlap, T.~Wu, B.~Zhu, J.~E. Gonzalez, and I.~Stoica.
\newblock From crowdsourced data to high-quality benchmarks: Arena-hard and benchbuilder pipeline.
\newblock \emph{arXiv preprint arXiv:2406.11939}, 2024.

\bibitem[Lin et~al.(2024{\natexlab{a}})Lin, Deng, Chandu, Brahman, Ravichander, Pyatkin, Dziri, Bras, and Choi]{lin2024wildbench}
B.~Y. Lin, Y.~Deng, K.~Chandu, F.~Brahman, A.~Ravichander, V.~Pyatkin, N.~Dziri, R.~L. Bras, and Y.~Choi.
\newblock Wildbench: Benchmarking llms with challenging tasks from real users in the wild.
\newblock \emph{arXiv preprint arXiv:2406.04770}, 2024{\natexlab{a}}.

\bibitem[Lin et~al.(2024{\natexlab{b}})Lin, Lin, Xiong, Diao, Liu, Zhang, Pan, Wang, Hu, Zhang, Dong, Pi, Zhao, Jiang, Ji, Yao, and Zhang]{lin2024mitigatingalignmenttaxrlhf}
Y.~Lin, H.~Lin, W.~Xiong, S.~Diao, J.~Liu, J.~Zhang, R.~Pan, H.~Wang, W.~Hu, H.~Zhang, H.~Dong, R.~Pi, H.~Zhao, N.~Jiang, H.~Ji, Y.~Yao, and T.~Zhang.
\newblock Mitigating the alignment tax of rlhf, 2024{\natexlab{b}}.
\newblock URL \url{https://arxiv.org/abs/2309.06256}.

\bibitem[Liu et~al.(2025)Liu, Mei, Lin, Xue, Wang, Xu, Wu, Zhang, Lin, Dong, et~al.]{liu2025deepseek}
A.~Liu, A.~Mei, B.~Lin, B.~Xue, B.~Wang, B.~Xu, B.~Wu, B.~Zhang, C.~Lin, C.~Dong, et~al.
\newblock Deepseek-v3. 2: Pushing the frontier of open large language models.
\newblock \emph{arXiv preprint arXiv:2512.02556}, 2025.

\bibitem[Liu et~al.(2024{\natexlab{a}})Liu, Zeng, Liu, Yan, He, Wang, Yan, Liu, and Zhou]{liu2024skywork}
C.~Y. Liu, L.~Zeng, J.~Liu, R.~Yan, J.~He, C.~Wang, S.~Yan, Y.~Liu, and Y.~Zhou.
\newblock Skywork-reward: Bag of tricks for reward modeling in llms.
\newblock \emph{arXiv preprint arXiv:2410.18451}, 2024{\natexlab{a}}.

\bibitem[Liu et~al.(2024{\natexlab{b}})Liu, Yao, Min, Cao, Hou, and Li]{liu2024rm}
Y.~Liu, Z.~Yao, R.~Min, Y.~Cao, L.~Hou, and J.~Li.
\newblock Rm-bench: Benchmarking reward models of language models with subtlety and style.
\newblock \emph{arXiv preprint arXiv:2410.16184}, 2024{\natexlab{b}}.

\bibitem[Mahan et~al.(2024)Mahan, Van~Phung, Rafailov, Blagden, Lile, Castricato, Fr{\"a}nken, Finn, and Albalak]{mahan2024generative}
D.~Mahan, D.~Van~Phung, R.~Rafailov, C.~Blagden, N.~Lile, L.~Castricato, J.-P. Fr{\"a}nken, C.~Finn, and A.~Albalak.
\newblock Generative reward models.
\newblock \emph{arXiv preprint arXiv:2410.12832}, 2024.

\bibitem[Malik et~al.(2025)Malik, Pyatkin, Land, Morrison, Smith, Hajishirzi, and Lambert]{malik2025rewardbench}
S.~Malik, V.~Pyatkin, S.~Land, J.~Morrison, N.~A. Smith, H.~Hajishirzi, and N.~Lambert.
\newblock Rewardbench 2: Advancing reward model evaluation.
\newblock \emph{arXiv preprint arXiv:2506.01937}, 2025.

\bibitem[{OpenAI}(2025)]{openai2025gpt5systemcard}
{OpenAI}.
\newblock {GPT-5 System Card}.
\newblock System card, OpenAI, Aug. 2025.
\newblock URL \url{https://cdn.openai.com/gpt-5-system-card.pdf}.
\newblock Accessed: 2026-01-05.

\bibitem[Ouyang et~al.(2022)Ouyang, Wu, Jiang, Almeida, Wainwright, Mishkin, Zhang, Agarwal, Slama, Ray, et~al.]{ouyang2022training}
L.~Ouyang, J.~Wu, X.~Jiang, D.~Almeida, C.~Wainwright, P.~Mishkin, C.~Zhang, S.~Agarwal, K.~Slama, A.~Ray, et~al.
\newblock Training language models to follow instructions with human feedback.
\newblock \emph{Advances in neural information processing systems}, 35:\penalty0 27730--27744, 2022.

\bibitem[Pyatkin et~al.(2025)Pyatkin, Malik, Graf, Ivison, Huang, Dasigi, Lambert, and Hajishirzi]{pyatkin2025generalizing}
V.~Pyatkin, S.~Malik, V.~Graf, H.~Ivison, S.~Huang, P.~Dasigi, N.~Lambert, and H.~Hajishirzi.
\newblock Generalizing verifiable instruction following.
\newblock \emph{arXiv preprint arXiv:2507.02833}, 2025.

\bibitem[Rafailov et~al.(2023)Rafailov, Sharma, Mitchell, Manning, Ermon, and Finn]{rafailov2023direct}
R.~Rafailov, A.~Sharma, E.~Mitchell, C.~D. Manning, S.~Ermon, and C.~Finn.
\newblock Direct preference optimization: Your language model is secretly a reward model.
\newblock \emph{Advances in neural information processing systems}, 36:\penalty0 53728--53741, 2023.

\bibitem[Rein et~al.(2024)Rein, Hou, Stickland, Petty, Pang, Dirani, Michael, and Bowman]{rein2024gpqa}
D.~Rein, B.~L. Hou, A.~C. Stickland, J.~Petty, R.~Y. Pang, J.~Dirani, J.~Michael, and S.~R. Bowman.
\newblock Gpqa: A graduate-level google-proof q\&a benchmark.
\newblock In \emph{First Conference on Language Modeling}, 2024.

\bibitem[Schulman et~al.(2017)Schulman, Wolski, Dhariwal, Radford, and Klimov]{schulman2017proximal}
J.~Schulman, F.~Wolski, P.~Dhariwal, A.~Radford, and O.~Klimov.
\newblock Proximal policy optimization algorithms.
\newblock \emph{arXiv preprint arXiv:1707.06347}, 2017.

\bibitem[Team et~al.(2025)Team, Bai, Bao, Chen, Chen, Chen, Chen, Chen, Chen, Chen, Chen, Cui, Ding, Dong, et~al.]{kimiteam2025kimik2}
K.~Team, Y.~Bai, Y.~Bao, G.~Chen, J.~Chen, N.~Chen, R.~Chen, Y.~Chen, Y.~Chen, Y.~Chen, Z.~Chen, J.~Cui, H.~Ding, M.~Dong, et~al.
\newblock Kimi k2: Open agentic intelligence.
\newblock \emph{arXiv preprint arXiv:2507.20534}, 2025.
\newblock URL \url{https://arxiv.org/abs/2507.20534}.

\bibitem[Tianle~Li(2024)]{stylearena2024}
W.-L.~C. Tianle~Li, Anastasios~Angelopoulos.
\newblock Does style matter? disentangling style and substance in chatbot arena, August 2024.
\newblock URL \url{https://blog.lmarena.ai/blog/2024/style-control/}.

\bibitem[Wang et~al.(2024)Wang, Dong, Delalleau, Zeng, Shen, Egert, Zhang, Sreedhar, and Kuchaiev]{wang2024helpsteer}
Z.~Wang, Y.~Dong, O.~Delalleau, J.~Zeng, G.~Shen, D.~Egert, J.~Zhang, M.~N. Sreedhar, and O.~Kuchaiev.
\newblock Helpsteer 2: Open-source dataset for training top-performing reward models.
\newblock \emph{Advances in Neural Information Processing Systems}, 37:\penalty0 1474--1501, 2024.

\bibitem[Wang et~al.(2025)Wang, Zeng, Delalleau, Shin, Soares, Bukharin, Evans, Dong, and Kuchaiev]{wang2025helpsteer3}
Z.~Wang, J.~Zeng, O.~Delalleau, H.-C. Shin, F.~Soares, A.~Bukharin, E.~Evans, Y.~Dong, and O.~Kuchaiev.
\newblock Helpsteer3-preference: Open human-annotated preference data across diverse tasks and languages.
\newblock \emph{arXiv preprint arXiv:2505.11475}, 2025.

\bibitem[Weng(2024)]{weng2024rewardhack}
L.~Weng.
\newblock Reward hacking in reinforcement learning.
\newblock \emph{lilianweng.github.io}, Nov 2024.
\newblock URL \url{https://lilianweng.github.io/posts/2024-11-28-reward-hacking/}.

\bibitem[Yang et~al.(2025)Yang, Li, Yang, Zhang, Hui, Zheng, Yu, Gao, Huang, Lv, et~al.]{yang2025qwen3}
A.~Yang, A.~Li, B.~Yang, B.~Zhang, B.~Hui, B.~Zheng, B.~Yu, C.~Gao, C.~Huang, C.~Lv, et~al.
\newblock Qwen3 technical report.
\newblock \emph{arXiv preprint arXiv:2505.09388}, 2025.

\bibitem[Zhang et~al.(2024)Zhang, Hosseini, Bansal, Kazemi, Kumar, and Agarwal]{zhang2024generative}
L.~Zhang, A.~Hosseini, H.~Bansal, M.~Kazemi, A.~Kumar, and R.~Agarwal.
\newblock Generative verifiers: Reward modeling as next-token prediction.
\newblock \emph{arXiv preprint arXiv:2408.15240}, 2024.

\bibitem[Zhang and Math-AI(2024)]{aime24}
Y.~Zhang and T.~Math-AI.
\newblock American invitational mathematics examination (aime) 2024, 2024.

\bibitem[Zhang and Math-AI(2025)]{aime25}
Y.~Zhang and T.~Math-AI.
\newblock American invitational mathematics examination (aime) 2025, 2025.

\bibitem[Zheng et~al.(2023)Zheng, Chiang, Sheng, Zhuang, Wu, Zhuang, Lin, Li, Li, Xing, et~al.]{zheng2023judging}
L.~Zheng, W.-L. Chiang, Y.~Sheng, S.~Zhuang, Z.~Wu, Y.~Zhuang, Z.~Lin, Z.~Li, D.~Li, E.~Xing, et~al.
\newblock Judging llm-as-a-judge with mt-bench and chatbot arena.
\newblock \emph{Advances in neural information processing systems}, 36:\penalty0 46595--46623, 2023.

\end{thebibliography}

\newpage
\appendix
\appendixpage  

\section{Limitations}

First, our Rationale-Centric Alignment (R-Align) framework relies on the availability of high-quality golden rationales to serve as the ground truth for logical verification. Ideally, these should be derived from dense human annotations, as seen in the HelpSteer3 dataset, which provides both preference labels and detailed critiques. However, such comprehensive human-annotated datasets are scarce. For the majority of training data, we were necessitated to employ advanced LLM to synthesize rationales conditioned on the ground-truth labels. This dependency implies that the efficacy of our method is currently bound by either the cost of human annotation or the reasoning capabilities of proprietary teacher models.

Second, we acknowledge the inherent subjectivity in evaluating reasoning: for a given preference pair, there may exist multiple valid logical paths that lead to the same conclusion. A rigid strict-matching approach could theoretically penalize valid but alternative reasoning. However, our empirical evaluation suggests this is less of a practical issue than a theoretical one. As shown in our benchmark results (Table ~\ref{tab:model_comparison}), advanced LLMs with strong reasoning capabilities (e.g., GPT-5-thinking, Gemini-2.5-Pro) exhibit very low rates of spurious correctness. This convergence indicates that for high-quality preference data, the underlying logic is sufficiently objective, and advanced LLMs tend to align consistently with the golden rationale, validating our reliance on rationale-based supervision.

\section{Data Construction Details}
\label{app:data_construction}

We construct our benchmark by augmenting existing datasets with high-quality reference critiques generated by {Gemini-3-Pro}. The processing details are as follows:

\textbf{Data Source \& Preprocessing.}
\begin{itemize}
    \item \textbf{RewardBench2}: The original dataset contains prompts with 4 responses (1 chosen, 3 rejected). We expand each entry into 3 independent pairwise samples by pairing the chosen response with each rejected response, assuming an Independent and Identically Distributed (I.I.D.) relationship. We explicitly exclude the ``Ties'' subset from our dataset construction, as it is specifically designed for scalar reward models.
    \item \textbf{PPE-Preference} \& \textbf{HelpSteer3}: We utilize the standard pairwise splits. For {HelpSteer3}, we collect all available attribute-specific human ratings and comments for each pair.
\end{itemize}

\textbf{Golden Judgment Generation.}
We employ {Gemini-3-Pro} to produce the reference critique $\mathbf{o}^*$ using two distinct strategies based on the availability of human annotations:
\begin{itemize}
    \item \textbf{Generation from Label (RewardBench2 \& PPE):} Since these datasets only provide the final preference label $l$, we prompt Gemini with the tuple $(x, y_1, y_2, l)$ to reverse-engineer the reasoning process. The model is instructed to justify why the preferred response is superior based on the ground truth. The full prompt used for this label-conditioned generation is presented in Figure \ref{app:rationale_prompt}.
    \item \textbf{Aggregation from Human Feedback (HelpSteer3):} This dataset includes multiple human judgments $\{h_1, h_2, h_3\}$. We feed the tuple $(x, y_1, y_2, \{h_i\}_{i=1}^3)$ to Gemini. The model acts as a ``Meta-Reviewer,'' synthesizing the diverse and potentially noisy human feedback into a single, comprehensive, and high-quality rationale $\mathbf{o}^*$. The comprehensive prompt employed for this meta-review aggregation is illustrated in Figure \ref{app:helpsteer_prompt}.
\end{itemize}

\begin{figure*}[t]
\centering
\begin{tcolorbox}[
    colback=red!5!white,          
    colframe=magenta!60!black,    
    coltitle=white,               
    title=\textbf{The Prompt for Label-Conditioned Rationale Generation},
    fonttitle=\small\bfseries,
    fontupper=\footnotesize\ttfamily, 
    boxrule=0.8pt,
    enhanced,
    breakable,
]
Act as an expert evaluator and analyst. You will be given a USER PROMPT, two AI assistant responses (A and B), and a **Ground Truth (GT) Preference Label** indicating which assistant provided the better response.\\
Your task is **not** to decide the winner yourself, but to **analyze and explain why the provided GT response is superior** based on the comparison.\\
You will compare A and B **relatively** without consulting external sources.\\
\#\# What to evaluate (to find justifications for the GT label)
Analyze the responses across these dimensions to identify why the GT response won:\\
1) **Factual accuracy \& correctness.** Check if the losing response contains errors that the GT response avoided.\\
2) **Instruction-following \& task completion.** Did the GT response follow instructions better?\\
3) **Relevance \& completeness.** Did the GT response cover more ground or stay more on-topic?\\
4) **Clarity \& concision.** Is the GT response better organized or more precise?\\
5) **Safety \& policy alignment.** Did the GT response handle safety better?\\
6) **Style \& Formatting.** If the content is similar, look for formatting or stylistic choices that make the GT response more readable.\\
\#\# Special Instructions for Justification\\
- **Support the Label:** Your analysis must conclude that the assistant specified by the GT label is the winner.\\
- **Find the Differentiator:** Focus on identifying the **decisive differences** that make the GT response better.\\
- **Materiality:** Prioritize material differences (accuracy, instructions). If no material differences exist, explain how minor factors (tone, brevity, formatting) justify the GT label.\\
- **Ambiguity:** If the User Prompt is ambiguous, explain how the GT response handled that ambiguity better (or why its interpretation was preferred).\\
- **Language:** Ensure the GT response followed language constraints appropriately.\\
\#\# How to structure your explanation\\
- **Requirements extracted from the USER PROMPT:** 2-5 bullet points.\\
- **Assistant A - strengths \& weaknesses:** 3-6 bullets. (Highlight traits that support the final verdict).\\
- **Assistant B - strengths \& weaknesses:** 3-6 bullets. (Highlight traits that support the final verdict).\\
- **Head-to-head comparison:** 2-4 bullets stating the decisive reasons why the GT assistant is better.\\
- **Missing but useful information (if any):** 1-3 bullets.\\
After your explanation, output the final verdict matches the GT label by wrapping the letter in \verb|\boxed{}|. Do not output any other text after the box.\\
Example:\\
\verb|\boxed{A}|\\
or\\
\verb|\boxed{B}|

\end{tcolorbox}
\vspace{-1.em}
\caption{The Prompt for Label-Conditioned Rationale Generation.
}
\label{app:rationale_prompt}
\end{figure*}

\section{Human Validation of Meta-Judging Reliability}
\label{app:human}
To validate the reliability of employing Gemini-3-Pro as the MetaRM for detecting rationale misalignment, we conducted a human agreement study. specifically, we sampled 53 instances from the HelpSteer3 benchmark where the Qwen3-14B model correctly predicted the preference label ($L\text{-}Acc = 1$).Human annotators were tasked with performing the meta-judge process: assessing whether the rationale generated by Qwen3-14B was logically consistent with the golden rationale. We then compared these human annotations against the judgments made by Gemini-3-Pro MetaRM using the prompt defined in Figure ~\ref{app:prompt_meta}. Treating the human judgments as the ground truth, Gemini-3-Pro achieved an F1 score of 0.9044. This high level of agreement confirms that our automated Meta-Judging pipeline serves as a reliable proxy for human evaluation in identifying spurious correctness.

\section{RLHF Implementation Details}
\label{app:rlhf_details}

In this section, we detail the specific configuration used for optimizing the policy models via Reinforcement Learning from Human Feedback (RLHF). To isolate the impact of the GenRM's supervision quality, all hyperparameters and configurations described below are kept identical across all experiments.

\paragraph{Policy Optimization.}
We initialize the policy model with the Qwen3-8B checkpoint. The optimization is performed using the Proximal Policy Optimization (PPO) algorithm~\citep{schulman2017proximal}.

\paragraph{Reference Response Generation.}
Since Generative Reward Models (GenRMs) typically operate by evaluating pairwise comparisons, a high-quality baseline is required for the policy to compete against during the training process. For each training prompt $x$, we generate a reference response $y_{\text{ref}}$ using the STEP3-VL-10B~\citep{huang2026step3} model. This ensures that the policy is constantly challenged by a strong upper-bound baseline.

\paragraph{Pairwise Judgment and Reward Formulation.}
During training, the GenRM functions as the judge for the generated outputs. For a given prompt, the GenRM compares the policy's sampled response $y_{\text{policy}}$ against the pre-generated reference $y_{\text{ref}}$. The reward signal $r$ is assigned based on the GenRM's verdict:
\begin{equation}
    r = 
    \begin{cases} 
    +1 & \text{if the GenRM judges } y_{\text{policy}} \succ y_{\text{ref}}, \\
    -1 & \text{otherwise}.
    \end{cases}
\end{equation}

\paragraph{Length Constraint.}
To mitigate length bias, we implement a strict, dynamic length penalty during the RL process. Specifically, the length constraint is determined by the relative difference between the generated response length $L_{gen}$ and the reference answer length $L_{ref}$, where $L_{gen}$ and $L_{ref}$ is calculated excluding the Chain-of-Thought content (i.e., ignoring content within \texttt{\textless think\textgreater} and \texttt{\textless /think\textgreater}).

A penalty is applied if the relative difference $\frac{L_{gen} - L_{ref}}{L_{ref}}$ exceeds a dynamic threshold $\delta(L_{ref})$. This threshold is not fixed; instead, it decays linearly based on the length of the reference answer:
\begin{equation}
    \delta(L_{ref}) = 
    \begin{cases} 
    0.60 & \text{if } L_{ref} \le 150 \\
    \text{Linear Interpolation} & \text{if } 150 < L_{ref} < 2000 \\
    0.40 & \text{if } L_{ref} \ge 2000
    \end{cases}
\end{equation}
This mechanism allows for more flexibility in shorter responses while enforcing stricter conciseness for longer outputs. If this constraint is violated, the response is marked as incorrect with a $-1$ reward.

\begin{figure*}[t]
\centering
\begin{tcolorbox}[
    colback=red!5!white,          
    colframe=magenta!60!black,    
    coltitle=white,               
    title=\textbf{The Meta-Reviewer Prompt for Rationale Aggregation},
    fonttitle=\small\bfseries,
    fontupper=\scriptsize\ttfamily\linespread{0.9}\selectfont, 
    boxrule=0.8pt,
    enhanced,
    breakable,
]

You are an expert AI evaluator acting as a "Meta-Reviewer." Your goal is to synthesize three separate expert analyses of two AI model responses into a single, authoritative judgment.\\
You must adopt the persona of a single, omniscient judge.\\ **Critically, you must treat the observations and findings provided in the "Expert Analyses" as the ground truth for your evaluation.** Your task is not to re-evaluate the models from scratch, but to articulate the consensus or strongest arguments found in the expert feedback as your own direct opinion.\\
\#\#\# Input Data Format\\
The user input will be structured using specific tags. You must parse the following sections:\\
1.  **Dialogue Context (Optional):** If present, the conversation history leading up to the final query will be enclosed in `<|Dialogue Context|>` tags.\\
2.  **User Prompt:** The current instruction or query to be evaluated is marked by `<|User Prompt|>`.\\
3.  **Assistant Responses:**\\
    * **Assistant A:** Content located between `<|The Start of Assistant A's Answer with User|>` and `<|The End of Assistant A's Answer with User|>`.\\
    * **Assistant B:** Content located between `<|The Start of Assistant B's Answer with User|>` and `<|The End of Assistant B's Answer with User|>`.\\
4.  **Expert Analyses:** A section following the responses containing the feedback from three experts (Note: Experts typically refer to A as "@Response 1" and B as "@Response 2").\\
\#\#\# Evaluation Criteria \& Consensus Handling\\
When synthesizing the expert feedback, apply the following logic:\\
1.  **Trust the Evidence:** If experts identify a hallucination, logic error, or safety risk, accept this as fact. Do not override expert findings based on your own internal knowledge unless the expert is blatantly violating the User Prompt.\\
2.  **Respect the Consensus:** If a majority of experts prefer one Assistant for specific reasons (e.g., better instruction following), your final verdict **must align with this preference**. Your job is to generate the *reasoning* that supports their conclusion, not to challenge it.\\
3.  **Resolve Disagreements (The Materiality Principle):** If experts disagree with each other:\\
    * Side with the expert pointing out **objective errors** (syntax, facts) over subjective preference (tone, style).
    * Side with the expert who strictly enforces the `<|User Prompt|>` constraints.\\
    * If one expert nitpicks minor wording while others praise the core logic, downweight the nitpick.\\
\#\#\# Guidelines\\
1.  **Unified Voice:** Do NOT mention "Expert 1" or "The reviewers." Write as if YOU analyzed the models directly (e.g., "Assistant A fails to..." instead of "The experts noted A fails to...").\\
2.  **Terminology:** Strictly refer to the models as **Assistant A** and **Assistant B**. (Map expert references of "@Response 1" to A and "@Response 2" to B).\\
3.  **Synthesis:** Do not just list points. Group them logically.\\
\#\#\# Output Structure\\
You must output your evaluation strictly following this format:\\
**Requirements extracted from the USER PROMPT:**\\
- {[Extract 2-5 key constraints/intentions from the User Prompt/Context, ensuring the models addressed them.]}\\
**Assistant A - strengths \& weaknesses:**\\
- {[Synthesize specific points from the experts regarding A. Use bullet points.]}\\
- {[Focus on factual correctness and instruction adherence as highlighted by the experts.]}\\
**Assistant B - strengths \& weaknesses:**\\
- {[Synthesize specific points from the experts regarding B. Use bullet points.]}\\
- {[Focus on factual correctness and instruction adherence as highlighted by the experts.]}\\
**Head-to-head comparison:**\\
- {[Synthesize the comparative arguments.]}\\
- {[Explain the decisive difference based on the expert consensus and Materiality Principle (e.g., "A is better because B has a logic error identified in the analysis").]}\\
**Missing but useful information (if any):**\\
- {[If experts noted anything missing, list it here. Otherwise, omit or say "None".]}\\
**Verdict:**\\
{[Conclude with the final verdict by wrapping the letter of the better assistant in \textbackslash boxed\{\}. Ensure this verdict aligns with the strongest objective arguments presented in the expert analyses.]}\\
Example:\\
\textbackslash boxed\{A\}

\end{tcolorbox}
\vspace{-1em}
\caption{The Meta-Reviewer Prompt for Rationale Aggregation.}
\label{app:helpsteer_prompt}
\end{figure*}

\begin{figure*}[t]
\centering
\begin{tcolorbox}[
    colback=red!5!white,          
    colframe=magenta!60!black,    
    coltitle=white,               
    title=\textbf{The MetaRM Prompt for Rationale Consistency Verification},
    fonttitle=\small\bfseries,
    fontupper=\footnotesize\ttfamily, 
    boxrule=0.8pt,
    enhanced,
    breakable,
]
\# Role\\
You are a professional RLHF data quality evaluation expert. Your task is to assess whether the "evaluation rationale" generated by a Reward Model (GenRM) accurately captures the core reasoning of a human expert (Golden Judge).\\
\# Input Data\\
Below are the conversation context and the responses from two models:\\
{context\_and\_responses}\\
Below is the evaluation provided by the human expert (Golden Judge):\\
<golden\_judge>\\
{golden\_explanation}\\
</golden\_judge>\\
Below is the evaluation generated by the model under review (GenRM):\\
<genrm\_output>\\
{genrm\_explanation}\\
</genrm\_output>\\
\# Evaluation Steps (Chain of Thought)\\
Please proceed step by step with the following analysis:\\
1. **Extract Golden Key Points**:\\
   Read the explanation in <golden\_judge> and identify the **core decisive factors** (Key Discriminators) that led to the final judgment (e.g., A > B).\\
   * Was it a factual error (hallucination)?\\
   * Was it a failure in instruction following?\\
   * Was it an issue of tone, formatting, or safety?\\
   * Note: Ignore generic politeness or boilerplate comments. Focus only on the specific logic that differentiates the quality of A and B.\\
2. **Check GenRM Coverage**:\\
   Read the explanation in <genrm\_output> and determine whether it **explicitly identifies** the above "core decisive factors."\\
   * If Golden says "A is wrong due to a math error," but GenRM says "A is wrong due to poor tone," even if both ultimately choose B as better, this is **Incorrect** (because the reasoning does not align and may be a lucky guess).\\
   * If GenRM only provides vague statements (e.g., "A is more detailed than B") without pointing out the specific issues emphasized by Golden, this is also **Incorrect**.\\
3. **Final Decision**:\\
   * If GenRM’s reasoning is consistent with Golden’s core logic (even if phrased differently), the verdict is **Correct**.\\
   * If GenRM misses key error points, fabricates reasons that do not exist, or conflicts with Golden’s logic, the verdict is **Incorrect**.\\
\# Output Format\\
Please strictly follow the XML format below when outputting your analysis and final conclusion:\\
<golden\_key\_points>\\
Briefly summarize the key points that the Golden Judge considers critical in distinguishing A from B\\
</golden\_key\_points>\\
<genrm\_analysis>\\
Analyze whether GenRM mentioned the above key points\\
</genrm\_analysis>\\
<final\_verdict>\\
Correct OR Incorrect\\
</final\_verdict>

\end{tcolorbox}
\vspace{-1.em}
\caption{The MetaRM Prompt for Rationale Consistency Verification.}
\label{app:prompt_meta}
\end{figure*}

\setcounter{figure}{0}
\makeatletter
\renewcommand{\thefigure}{A\@arabic\c@figure}
\makeatother

\setcounter{table}{0}
\makeatletter
\renewcommand{\thetable}{A\@arabic\c@table}
\makeatother

\end{document}